\documentclass[twocolumn,10pt]{article}
\usepackage[linespread=1.2]{kola}

\usepackage{graphicx}
\usepackage{subcaption}
\usepackage{stfloats}
\usepackage{comment}
\usepackage[fixed]{fontawesome6}
\usepackage{amsmath}
\usepackage{amssymb}

\usepackage{graphicx}

\usepackage{booktabs}

\usepackage[table]{xcolor}

\usepackage{nicematrix}
\usepackage{placeins}
\usepackage{float}
\usepackage{tikz}
\usetikzlibrary{patterns, patterns.meta}

\newcommand{\dashmidrule}{%
  \noalign{\vskip\aboverulesep}%
  \Hline[tikz={dash pattern=on 2pt off 5pt}]%
  \noalign{\vskip\belowrulesep}%
}

\usepackage{makecell}

\usepackage{pifont}

\definecolor{ArchivalGray}{HTML}{cccccc}
\kolarulecolor{ArchivalGray}
\kolaheadertext{}

\title{PreFIQs: Face Image Quality Is What Survives Pruning}

\author{Jan Niklas Kolf \textsuperscript{\textrm 1,\textrm 3} \;
Guray Ozgur \textsuperscript{\textrm 1} \;
Andrea Atzori \textsuperscript{\textrm 1} \;
Žiga Babnik \textsuperscript{\textrm 2} \\
Vitomir Štruc \textsuperscript{\textrm 2} \; 
Naser Damer \textsuperscript{\textrm 1,\textrm 3} \;
Fadi Boutros \textsuperscript{\textrm 1}}

\affiliation{%
    \textsuperscript{\rm 1} Fraunhofer Institute for Computer Graphics Research IGD, Germany\\
    \textsuperscript{\rm 2} University of Ljubljana, Slovenia\\
    \textsuperscript{\rm 3} Technical University of Darmstadt, Germany
}

\addkolalink{\faGithub github.com/jankolf/PreFIQs}{https://github.com/jankolf/PreFIQs}
\addkolalink{\faEnvelope jan.niklas.kolf@igd.fraunhofer.de}{mailto:jan.niklas.kolf@igd.fraunhofer.de}

\kolaabstracttext{%
Face Image Quality Assessment (FIQA) evaluates the utility of a face image for automated face recognition (FR) systems. In this work, we propose PreFIQs, an unsupervised and training-free FIQA framework grounded in the Pruning Identified Exemplar (PIE) hypothesis. We hypothesize that low-utility face images rely disproportionately on fragile network parameters, resulting in larger geometric displacement of their embeddings under model sparsification. Accordingly, PreFIQs quantifies image utility as the Euclidean distance between L2-normalized embeddings extracted from a pre-trained FR model and its pruned counterpart. 
We provide a first-order theoretical justification via a Jacobian-vector product analysis, demonstrating that this empirical drift serves as a computationally efficient approximation of the exact geometric sensitivity of the latent embedding manifold. Extensive experiments across eight benchmarks and four FR models demonstrate that PreFIQs achieves competitive or superior performance compared to state-of-the-art FIQA methods, including establishing new state-of-the-art results on several benchmarks, without any training or supervision. These results validate parameter sparsification as a principled and practically efficient signal for face image utility, and demonstrate that quality 
is, in essence, what survives pruning. 
}
\kolakeywords{Face Image Quality Assessment,\\Face Recognition,\\Model Pruning}
\kolavenue{IEEE/CVF Conference on\\Computer Vision and Pattern\\Recognition 2026 --\\Biometrics Workshop}
\kolalogo[width=5.0cm]{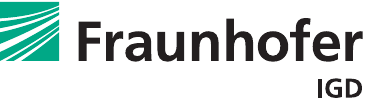}

\begin{document}
\kolamaketitle

\section{Introduction}
\label{sec:introduction}

Face Recognition (FR) systems have achieved remarkable accuracy \cite{DBLP:conf/cvpr/Kim0L22,deng2019arcface,wang2018cosfacelargemargincosine}, yet their performance can degrade in unconstrained and challenging real-world settings. 
Images captured in the wild often exhibit extreme variations in pose, illumination, occlusion, and blur, posing significant challenges for recognition \cite{Zhao2018PIM}. 
To address these issues, Face Image Quality Assessment (FIQA) has emerged as a critical preprocessing step, measuring the utility of a face image for automated recognition \cite{ISOIEC29794-5}. High-utility images produce stable and discriminative embeddings leading to reliable recognition, while low-utility images generate uncertain embeddings, undermining the robustness of FR systems~\cite{SERFIQ, BiyingWACV,PFE_FIQA}.

Based on the type of supervision, current state-of-the-art (SOTA) FIQA methods can be broadly categorized as supervised, unsupervised, or self-supervised. Supervised approaches \cite{faceqnetv1,SDDFIQA,RANKIQ_FIQA} rely on explicit or proxy labels to learn quality scores. Self-supervised approaches \cite{boutros_2023_crfiqa,MagFace,PFE_FIQA,atzori2025vitfiqaassessingfaceimage,Ou_2024_CVPR} jointly optimize FR and FIQA. Unsupervised approaches \cite{SERFIQ,grafiqs,FaceQAN,vitnt_fiqa,FROQ}, including our proposed PreFIQs, infer quality by evaluating the robustness of embeddings from pre-trained FR models under stochastic perturbations.
A central hypothesis in unsupervised FIQA is that high-utility images produce representations that are resilient to perturbations. For example, SER-FIQ~\cite{SERFIQ} measures embedding variance across multiple forward passes with random dropout, DifFIQA~\cite{10449044} leverages diffusion processes to quantify robustness against noise, and ViTNT-FIQA~\cite{vitnt_fiqa} tracks the stability of feature evolution across transformer blocks. While effective, stochastic methods like SER-FIQ incur substantial computational overhead due to repeated inference, and gradient-based methods such as GraFIQs~\cite{grafiqs} require backpropagation, which can be prohibitively expensive for real-time deployment.

In this work, we propose \textbf{PreFIQs}, a training-free FIQA framework that measures image utility through model sparsity sensitivity. Our method is grounded in the observation that high-utility images produce feature representations that remain stable under moderate network pruning, whereas low-utility images rely on fragile, easily disrupted parameters. Concretely, we compute the Euclidean distance between L2-normalized embeddings generated by a pre-trained FR model and its pruned counterpart. This embedding drift serves as a proxy for image utility: smaller drift indicates stable identity encoding and high quality, while larger drift signals sensitivity to pruning and lower utility.
By interpreting embedding stability under model sparsity as a quality measure, PreFIQs offers a new principled perspective on image utility. The method is fully training-free, requires no labels, and directly captures the functional contribution of each image to the recognition model's robustness. We validate PreFIQs across seven standard benchmarks and four FR models, demonstrating competitive or superior performance compared to SOTA supervised and unsupervised FIQA methods.

\vspace{-1mm}
\section{Related Work}
\label{sec:related_work}
\vspace{-1mm}
Face Image Quality Assessment (FIQA) methods have evolved along several complementary directions, which can be broadly categorized into three paradigms: supervised, unsupervised, and self-supervised approaches.  

\textbf{Supervised approaches} typically train quality regressors using explicit or proxy supervision. For example, FaceQnet~\cite{faceqnetv1} relies on ICAO compliance labels, SDD-FIQA~\cite{SDDFIQA} derives pseudo-labels from similarity-distribution distances, and RankIQ~\cite{RANKIQ_FIQA} formulates FIQA as a learning-to-rank problem. Subsequent works improve the reliability of these labels: CLIB-FIQA~\cite{10656894} calibrates the confidence of quality anchors, while MR-FIQA~\cite{MRFIQA} leverages multi-reference representations generated from synthetic data to reduce label noise.

\textbf{Unsupervised approaches} can be further divided into non-FR model approaches and FR-specific approaches. Non-FR model approaches estimate face quality without relying on conventional FIQA regressors or pretrained FR. DifFIQA~\cite{10449044} measures sample robustness through diffusion-based modeling, and eDifFIQA~\cite{babnikTBIOM2024} distills this into a lightweight predictor. DSL-FIQA~\cite{10657603} combines degradation-aware representation learning with landmark-guided transformers. %

FR-specific approaches probe frozen FR backbones without retraining to estimate FIQ. SER-FIQ~\cite{SERFIQ} measures embedding stability under dropout perturbations, GraFIQs~\cite{grafiqs} exploits gradient-based signals, and FaceQAN~\cite{FaceQAN} links quality to adversarial robustness. Recent training-free methods extend these ideas to transformer architectures and intermediate layers: ViTNT-FIQA~\cite{ozgur2026vitntfiqatrainingfreefaceimage} tracks embedding-trajectory stability across ViT layers, while FROQ~\cite{FROQ} identifies informative intermediate layers via a lightweight calibration step to predict quality in a single forward pass. These methods represent a shift toward efficient, probe-based FIQA without additional supervision.  

\textbf{Self-supervised approaches}, often implemented as FR-integrated methods, jointly optimize FR and FIQA. MagFace~\cite{MagFace} links quality to embedding magnitude, PFE~\cite{PFE_FIQA} models uncertainty in embeddings as a quality proxy, and ViT-FIQA~\cite{atzori2025vitfiqaassessingfaceimage} introduces a learnable quality token that directly predicts FIQ scores, while CR-FIQA~\cite{boutros_2023_crfiqa} explicitly learns \emph{relative classifiability} across identities, providing a task-relevant measure of utility rather than relying on surrogate labels or embedding magnitude.  

Building on these insights, \textbf{PreFIQs} introduces a complementary perspective: it quantifies image utility through \emph{sparsity-induced representational drift}. By measuring how sparsifying model parameters affects the embeddings of each image, PreFIQs captures the functional importance of samples for the recognition model itself. This training-free, data-free metric with minimal computational overhead provides a deterministic proxy for utility, emphasizing model robustness and discriminative stability, distinguishing it from prior FIQA approaches.

\vspace{-1mm}
\section{Methodology}
\label{sec:methodology}
\vspace{-1mm}
In this section, we introduce Pruning-based Face Image Quality Assessment (PreFIQs). PreFIQs quantifies the utility of face images by measuring the representational drift induced by controlled model sparsification. Our method builds on the Pruning Identified Exemplar (PIE) hypothesis~\cite{Hooker2019WhatDoCompressed}, which proves that the performance of compressed Deep Neural Networks (DNNs) disproportionately degrades on difficult or low-quality samples. We extend this principle to FR and hypothesize that low-utility face images exhibit higher sensitivity to parameter pruning, resulting in larger geometric displacement in the embedding space. Conversely, high-utility samples produce identity representations that remain stable under moderate structural compression.

Leveraging this asymmetry, we define face image quality as the stability of L2-normalized embeddings under pruning, measured as the Euclidean distance between embeddings extracted from the original and sparsified models. This drift-based formulation provides a deterministic and architecture-aligned proxy for face image utility, requiring neither additional training nor auxiliary supervision.

\subsection{Preliminary on Model Pruning}
Model pruning is a form of model compression that reduces the effective capacity of a DNN by removing redundant parameters~\cite{pruning_survey, 10.5555/3546258.3546499}. Recent SOTA FR models are typically over-parameterized \cite{pruning_survey, ALONSOFERNANDEZ2025221}, allowing substantial parameter removal while maintaining strong verification performance.
Pruning strategies can be broadly categorized along two principal dimensions: (i) the \emph{pruning criterion} used to identify removable parameters, and (ii) the \emph{pruning granularity}, i.e., whether parameters are removed individually (unstructured) or in groups (structured).

Let an FR model be denoted by 
$M_{\theta} : x \rightarrow \mathbb{R}^d,$ that map input $x$ to $\mathbb{R}^d$ embedding and parameterized by $\theta \in \mathbb{R}^{N}$, where $N$ denotes the total number of learnable parameters. Given a target sparsity ratio $\rho \in (0,1)$, pruning aims to construct a sparsified parameter vector $\theta_{\rho}$ satisfying:
\begin{equation}
\|\theta_{\rho}\|_0 = (1-\rho)N,
\end{equation}
where $\|\cdot\|_0$ denotes the $\ell_0$ pseudo-norm counting non-zero entries.

Pruning can be formulated as the application of a binary mask $\mathbf{m}_{\rho} \in \{0,1\}^{N}$ to the original parameters:

\begin{equation}
\label{eq:mask}
\theta_{\rho} = \mathbf{m}_{\rho} \odot \theta,
\end{equation}

where $\odot$ denotes the element-wise product. The mask $\mathbf{m}_{\rho}$ is constructed such that a fraction $\rho$ of parameters is set to zero.

\textbf{Pruning criterion.}
The pruning criterion determines how parameters are selected for removal. As a baseline, parameters can be removed uniformly at random, independent of their magnitude or functional contribution. However, random pruning does not explicitly target redundant parameters and often leads to lower accuracy compared to methods that remove unimportant ones \cite{10.5555/3495724.3496259, DBLP:conf/iclr/FrankleD0C21}. Importance can also be estimated using first-order information, e.g., the magnitude of gradients with respect to a loss function $\mathcal{L}$. Parameters with small $|\partial \mathcal{L} / \partial \theta_i|$ are considered less influential and can be pruned. However, this criterion required access to a training dataset to select parameters to be pruned, which is out of scope of this work, where we propose a training- and data-free FIQA approach.  
A common and effective strategy is magnitude-based pruning, where parameters with the smallest absolute values are removed under the assumption that low-magnitude weights contribute less to the model output.  In this case, a threshold $\tau$ is determined such that

\begin{equation}
m_{\rho,i} = \mathbb{I}(|\theta_i| > \tau),
\end{equation}

where $\mathbb{I}(\cdot)$ denotes the indicator function. This selection mechanism assumes that parameters with small magnitude contribute less to the network output and can therefore be removed with limited impact on global performance.

\textbf{Granularity of Pruning.} Pruning can be applied either in an unstructured manner, where individual weights are set to zero while preserving the network topology, or in a structured manner, where entire parameter groups (e.g., filters or channels) are removed, requiring corresponding architectural adjustments.

In PreFIQs, pruning is not used for computational acceleration but as a controlled mechanism to systematically reduce model capacity. The sparsified model $M_{\theta_{\rho}}$ therefore provides a principled means of analyzing how embedding representations respond to reductions in network capacity.

\begin{figure*}[!t]
\centering
\includegraphics[width=0.8\linewidth, trim=1 1 1 1,clip]{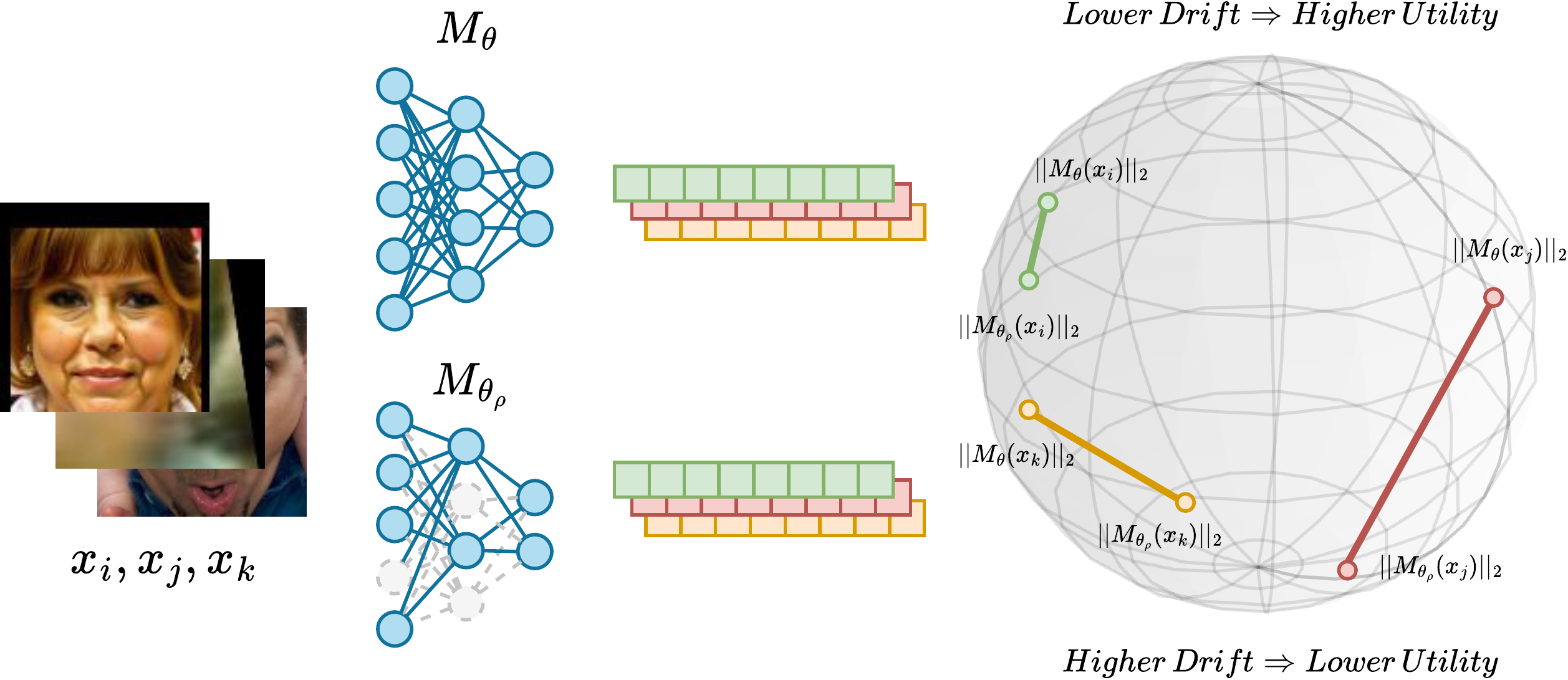}
%\vspace{-5mm}
\caption{Given face images, e.g, $x_i$, $x_j$, and $x_k$, we extract their L2-normalized embeddings using a pre-trained FR and its sparsified counterpart.  FIQ is quantified, for each image, as the Euclidean distance between its corresponding embeddings, measuring the pruning-induced representation drift. Smaller drift indicates stable identity encoding and thus higher image utility, while larger drift reflects structural sensitivity and lower quality.}
\label{fig:pipeline}
%\vspace{-4mm}
\end{figure*}

\subsection{PreFIQs}
Recent FR models \cite{liu2017sphereface, deng2019arcface, wang2018cosfacelargemargincosine} encode identity information in the angular direction of the embedding space. Consequently, feature representations are L2-normalized and lie on the unit hypersphere. Let $
M_{\theta}(x) \in \mathbb{R}^d
\quad \text{and} \quad
M_{\theta_{\rho}}(x) \in \mathbb{R}^d $ denote the L2-normalized embeddings of an input sample $x \in \mathcal{X}$ extracted by the original and sparsified FR models, respectively.

Building upon the PIE hypothesis \cite{Hooker2019WhatDoCompressed, jiang2021sdclr}, we interpret pruning as a controlled reduction of model capacity that exposes the structural dependence of a sample’s representation on specific parameters. If the identity encoding of $x$ relies heavily on parameters removed during pruning, its embedding will undergo a measurable geometric displacement. Conversely, embeddings that are encoded in more redundant or stable parameter subspaces will remain comparatively invariant under moderate sparsification.
We therefore quantify the utility of a sample $x$ by measuring the representation drift induced by pruning:
\begin{equation}
\label{eq:drift_calculation}
D(x) = \left\| M_{\theta}(x) - M_{\theta_{\rho}}(x) \right\|_2.
\end{equation}

Since both embeddings are L2-normalized, they lie on the unit hypersphere, and the Euclidean distance is bounded:
\begin{equation}
0 \leq D_{\rho}(x) \leq 2.
\end{equation}
Moreover, the Euclidean distance between normalized embeddings is directly related to angular deviation:
\begin{equation}
D^2(x) = 2 - 2 \cos\left(\angle \big(M_{\theta}(x), M_{\theta_{\rho}}(x)\big)\right),
\end{equation}
demonstrating that $D(x)$ measures the angular displacement of identity information in latent space.

To obtain a normalized FIQ score $Q(x) \in [0,1]$, where higher values indicate higher utility, we apply linear rescaling:
\begin{equation}
\label{eq:prefiqs_quality}
Q(x) = 1 - \frac{D(x)}{2}.
\end{equation}
Under this formulation:
$
Q(x) \approx 1
\quad \Leftrightarrow \quad
\text{high embedding stability (high utility)},$
and $
Q(x) \approx 0
\quad \Leftrightarrow \quad
\text{large structural sensitivity (low utility)}.$

Importantly, this drift-based formulation is deterministic, requires no auxiliary supervision or stochastic perturbations, and directly aligns the quality estimate with the geometry of the identity embedding manifold.

\begin{figure*}[!t]
    \centering

    \begin{subfigure}[t]{0.19\linewidth}
        \vspace{0pt} %
        \includegraphics[width=\linewidth]{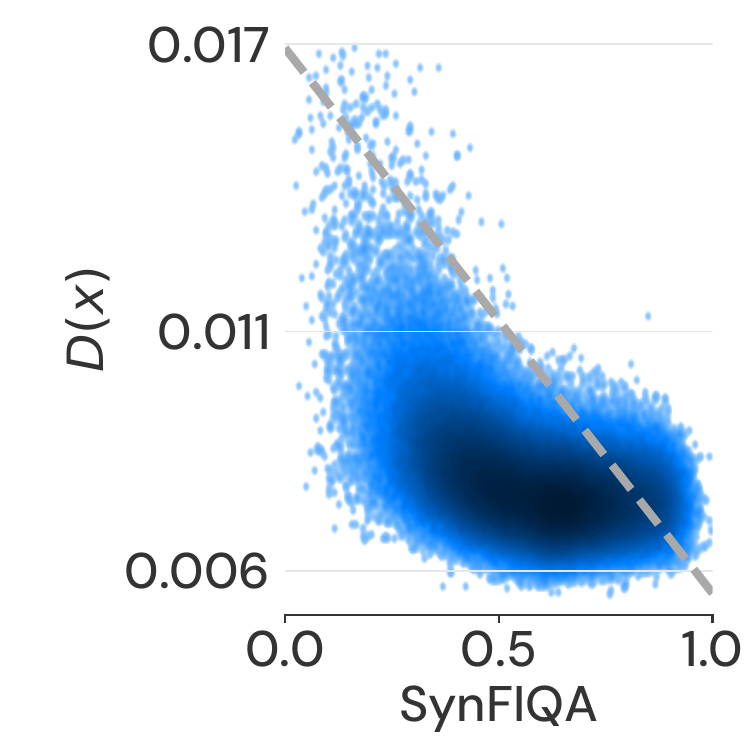}
        \caption{$D(x)$ (Eq.~\ref{eq:drift_calculation})}
        \label{fig:heatmap_pruned_estimate}
    \end{subfigure}
    \hfill
    \begin{subfigure}[t]{0.19\linewidth}
        \vspace{0pt} 
        \includegraphics[width=\linewidth]{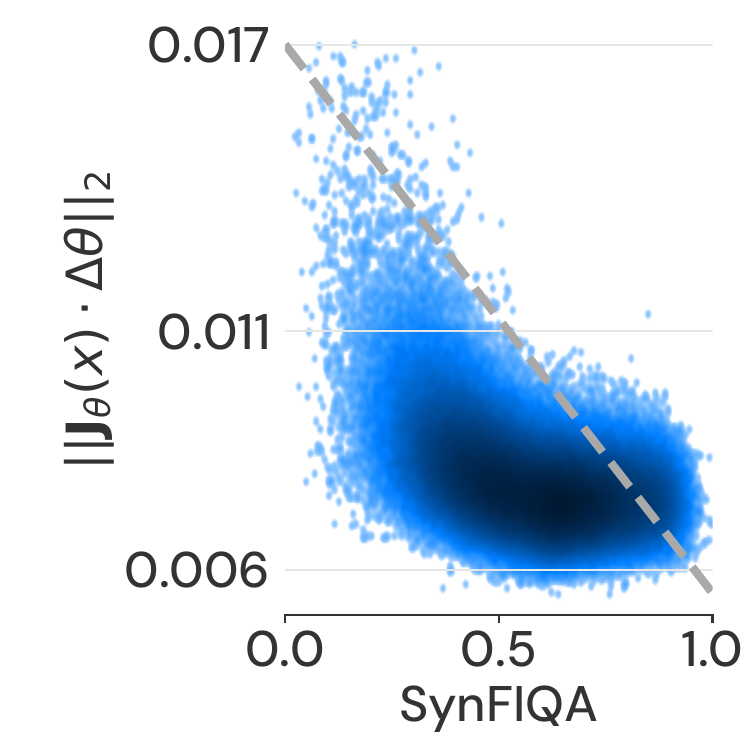}
        \caption{Jacobian Drift (Eq.~\ref{eq:jacobian_drift})}
        \label{fig:heatmap_jacobian_estimate}
    \end{subfigure}
    \hfill
    \begin{subfigure}[t]{0.19\linewidth}
        \vspace{0pt} 
        \includegraphics[width=\linewidth]{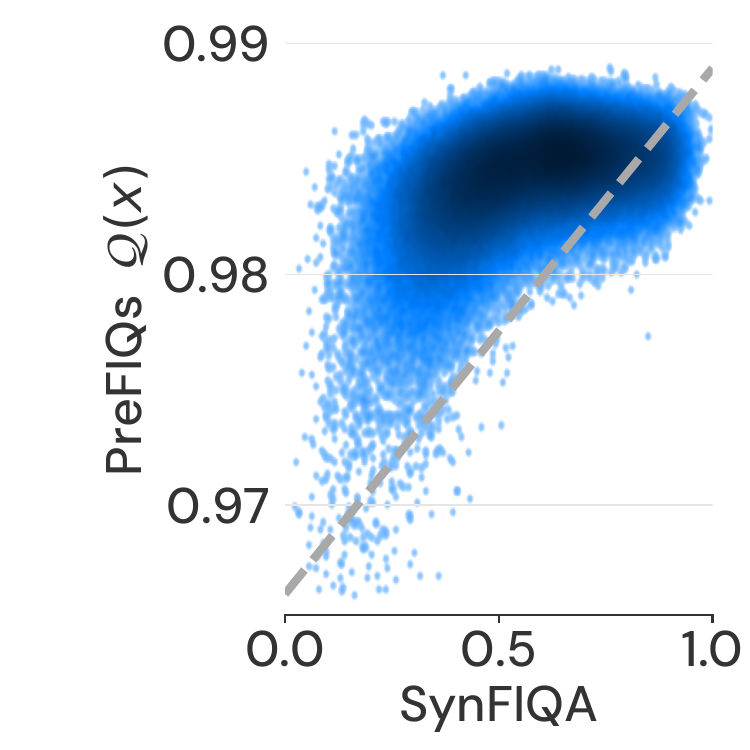}
        \caption{PreFIQs (Ours, Eq.~\ref{eq:prefiqs_quality})}
        \label{fig:heatmap_prefiqs}
    \end{subfigure}
    \hfill
    \begin{subfigure}[t]{0.19\linewidth}
        \vspace{0pt} 
        \includegraphics[width=\linewidth]{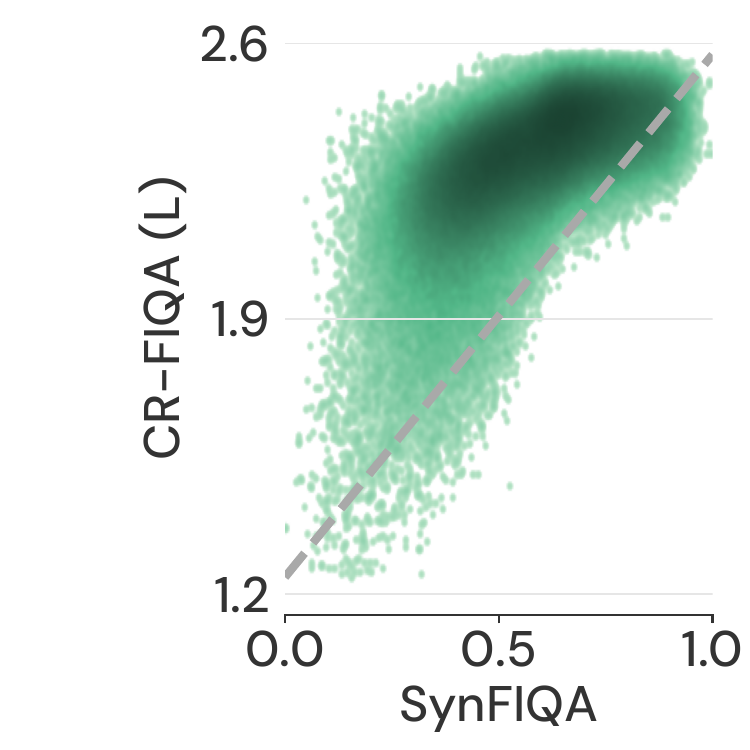}
        \caption{CR-FIQA (L) \cite{boutros_2023_crfiqa}}
        \label{fig:heatmap_crfiqa}
    \end{subfigure}
    \hfill
    \begin{subfigure}[t]{0.19\linewidth}
        \vspace{0pt} 
        \includegraphics[width=\linewidth]{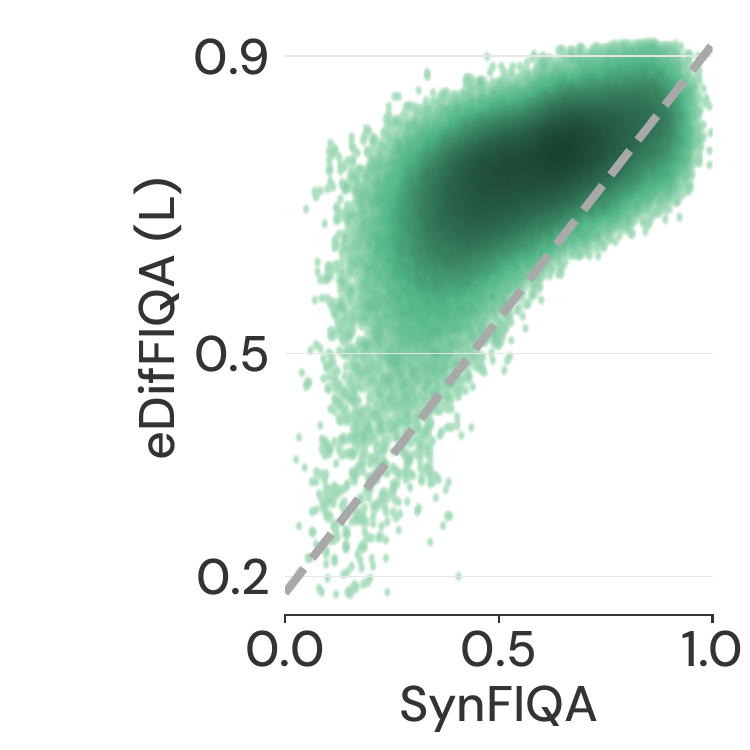}
        \caption{eDifFIQA (L) \cite{babnikTBIOM2024}}
        \label{fig:heatmap_ediffiqa}
    \end{subfigure}
    \vspace{-2mm}
    
    \caption{Density maps using SynFIQA~\cite{MRFIQA} dataset (550k images), and their proxy labels (x-axis, higher value indicates higher utility) versus various FIQA predictions (y-axis). Figs.~\ref{fig:heatmap_pruned_estimate} and \ref{fig:heatmap_jacobian_estimate} validate our approximation (Eq.~\ref{eq:jacobian_drift}), showing consistent distributions between Jacobian-based drift (Eq.~\ref{eq:jacobian_drift}) and empirical pruned-model distance (Eq.~\ref{eq:drift_calculation}, lower indicates higher utility). Figs.~\ref{fig:heatmap_prefiqs}--\ref{fig:heatmap_ediffiqa} compare our normalized, unsupervised PreFIQs score with supervised CR-FIQA~\cite{boutros_2023_crfiqa} and eDifFIQA~\cite{babnikTBIOM2024} (higher value indicates higher utility). Note that SynFIQA labels are algorithmic pseudo-labels, inherently biased by the used synthetic generation, rather than absolute ground truth.}
    \label{fig:total_heatmap_comparison}
    \vspace{-5mm}

\end{figure*}

\subsubsection{Theoretical Validation via Jacobian-Vector Product}
\label{sec:methodology_subsubseec:theoretical_validation}

To mathematically validate why the empirical drift $D(x)$ serves as a principled proxy for face image utility, we formalize the model's sensitivity to sparsification using a first-order Taylor expansion. We model the pruning process as an additive structural perturbation $\Delta\theta$ applied to the network weights, where $\Delta\theta_i = -\theta_i$ if weight $\theta_i$ is pruned, and $0$ otherwise, yielding $\theta_\rho = \theta + \Delta\theta$. Note that this additive formulation is equivalent to the mask-based sparsification in Eq.~\ref{eq:mask}, where 
$\Delta\theta_i = -\theta_i \cdot (1 - m_{\rho,i})$.
Under moderate sparsification, where $\|\Delta\theta\|$ remains sufficiently small, the perturbed face embedding can be approximated as:
\begin{equation}
\label{eq:taylor_expansion}
    M_{\theta + \Delta\theta}(x) \approx M_{\theta}(x) 
    + \mathbf{J}_{\theta}(x) \cdot \Delta\theta,
\end{equation}
where $\mathbf{J}_{\theta}(x) \in \mathbb{R}^{d \times N}$ is the Jacobian of the 
L2-normalized embedding with respect to the weights $\theta$, evaluated at input $x$.
Rearranging Eq.~\ref{eq:taylor_expansion} and taking the $\ell_2$-norm of both sides, the magnitude of the theoretical representation drift is governed by the norm of the Jacobian-vector product:
\begin{equation}
\label{eq:jacobian_drift}
    ||\mathbf{J}_{\theta}(x) \cdot \Delta\theta||_2 
    \approx ||M_{\theta + \Delta\theta}(x) - M_{\theta}(x)||_2.
\end{equation}

Recent work~\cite{grafiqs} validates gradient magnitudes as a robust indicator of FIQ, where high-utility images induce low gradient magnitudes, while low-utility samples require parameter updates of higher magnitude to resolve the distribution shift measured by an auxiliary loss. However, rather than depending on backpropagation of an auxiliary distribution shift loss~\cite{grafiqs}, PreFIQs directly probes the geometric sensitivity of the latent face embedding manifold via $\mathbf{J}_{\theta}(x)$, approximated without any gradient computation through the 
forward pass of the sparsified model.

Under our hypothesis that sample utility governs reliance on specific parameter subspaces, the following asymmetry is expected. Let $x_\text{high}$ be a sample with substantially higher utility than $x_\text{low}$. Since $x_\text{high}$ encodes identity in redundant, distributed parameter subspaces, it is comparatively robust to 
sparsification, and $\Delta\theta$ is expected to isolate near-zero Jacobian entries, yielding $||\mathbf{J}_{\theta}(x_\text{high}) \cdot \Delta\theta||_2 \approx 0$. Conversely, $x_\text{low}$ relies more heavily on the pruned weights in $\Delta\theta$, 
yielding Jacobian entries of higher magnitude and consequently a stronger drift:
\begin{equation}
    ||\mathbf{J}_{\theta}(x_\text{low}) \cdot \Delta\theta||_2 
    \gg ||\mathbf{J}_{\theta}(x_\text{high}) \cdot \Delta\theta||_2.
\end{equation}

While the Jacobian-vector product $||\mathbf{J}_{\theta}(x) \cdot \Delta\theta||_2$ exactly models this structural sensitivity, explicitly computing the full Jacobian $\mathbf{J}_{\theta}(x)$ is computationally intractable for SOTA architectures with 
tens of millions of parameters. Even forward-mode automatic differentiation introduces significant overhead. Equation~\ref{eq:jacobian_drift} establishes that our proposed 
empirical distance $D(x)$ (Eq.~\ref{eq:drift_calculation}) is a first-order approximation of this exact geometric sensitivity, providing a computationally efficient surrogate that requires neither backpropagation nor an auxiliary distribution shift loss~\cite{grafiqs}.

To validate this theoretical derivation, we analyze the representation drift on SynFIQA~\cite{MRFIQA}, a comprehensive synthetic dataset constructed to systematically model diverse intra-class quality degradations. As shown qualitatively in Figures~\ref{fig:heatmap_jacobian_estimate} and~\ref{fig:heatmap_pruned_estimate}, the density distributions of the exact Jacobian-vector product ($||\mathbf{J}_{\theta}(x) \cdot \Delta\theta||_2$) and the empirical Euclidean distance ($D(x)$) align almost perfectly across the quality spectrum, confirming that the static pruned model accurately captures the geometric sensitivity of the latent manifold. Quantitative validation on standard evaluation benchmarks is provided in Section~\ref{sec:results}.

Furthermore, Figures~\ref{fig:heatmap_prefiqs} through~\ref{fig:heatmap_ediffiqa} 
contrast our final PreFIQs score ($Q(x)$) against the proxy labels of the 
SynFIQA. Notably, our training-free PreFIQs yields a quality 
density distribution closely aligned with SOTA supervised methods, such as 
CR-FIQA~\cite{boutros_2023_crfiqa} and eDifFIQA~\cite{babnikTBIOM2024}, which 
require explicit training phases to learn quality regression.

\section{Experimental Setup}
\label{sec:experimental_setup}

\textbf{Pretrained Model Architecture.}
We demonstrate the proposed PreFIQs approach using two publicly available pre-trained FR models released by \cite{deng2019arcface}. Specifically, we utilize a ResNet100 model trained on MS1MV2 \cite{guo_2016_ms1m, deng2019arcface} and a ResNet50 model trained on CASIA-WebFace \cite{casia_webface}, both optimized using the ArcFace loss function. %

\textbf{Validating the Jacobian Approximation.}
We verify the accuracy of our proposed distance metric by comparing it directly to the exact mathematical formula (Jacobian-Vector product). To do this, we calculate both the exact gradient-based drift and our simpler Euclidean distance $\mathcal{D}(x)$ across the datasets. The exact Jacobian product is computed using PyTorch's \cite{NEURIPS2019_9015} automatic differentiation tools. We perform this comparison by pruning 10\% of the model's weights ($\rho = 0.1$). Finally, we measure how closely the two methods align using the pAUC score up to a 30\% discard rate.

\textbf{Model Pruning.}
We implement a global pruning strategy that includes all parameters within the convolutional and linear layers of the evaluated architectures.
To systematically assess the impact of parameter reduction, we perform a comparative analysis across the granularity of pruning unstructured and structured as well as pruning criterion and random pruning (baseline) and magnitude-based pruning.
Unstructured pruning is using $L_1$-norm based magnitude pruning, and the ratio of $\rho$ lowest magnitude parameters are pruned.
Structured pruning is performed using the DepGraph framework \cite{Fang2023DepGraphTA} to manage architectural dependencies, and the final linear layer is not pruned in structural pruning to achieve the same face embedding dimensionality.
Random pruning is randomly selecting parameters to prune to match pruning ratio $\rho$.
The models are pruned across a comprehensive spectrum of target sparsity ratios, defined as $\rho \in \{0.1, 0.2, 0.3, 0.4, 0.5, 0.6, 0.7, 0.8, 0.9\}$.

\textbf{FR Performance.}
To assess the impact of model pruning on the FR verification performance, we evaluate the pruned FR models on a set of diverse evaluation benchmarks, including Labeled Faces in the Wild (LFW) \cite{LFWTech}, AgeDB (using 30 year gap protocol) \cite{agedb}, Celebrities in Frontal-Profile in the Wild (CFP-FP) \cite{cfp-fp}, Cross-Age LFW (CALFW) \cite{CALFW}, and Cross-Pose LFW (CPLFW) \cite{CPLFWTech} using their official evaluation protocols..

\textbf{Evaluation Benchmarks.}
To ensure alignment with recent SOTA FIQA evaluation protocols \cite{boutros_2023_crfiqa}, we report our results across seven standard benchmarks: LFW\cite{LFWTech}, AgeDB  \cite{agedb}, CFP-FP \cite{cfp-fp}, CALFW \cite{CALFW}, Adience \cite{Adience}, CPLFW \cite{CPLFWTech}, Cross-Quality LFW (XQLFW) \cite{XQLFW} and IARPA Janus Benchmark–C (IJB-C) \cite{ijbc}. These datasets introduce a diverse set of challenging verification scenarios, containing significant variations in age (AgeDB and CALFW), head-pose (CFP-FP and CPLFW) and overall image quality (XQLFW).

\textbf{Evaluation Metrics.}
We assess FIQA performance utilizing Error-Versus-Discard Characteristic (EDC) curves \cite{GT07,NISTQuaity}, a standard evaluation metric in the literature (often referred to interchangeably as Error-Versus-Reject Curves, or ERC \cite{10449044}). The EDC curve illustrates the impact of sequentially discarding a fraction of the lowest-quality face images on the overall face verification performance. This performance is measured by the False Non-Match Rate (FNMR) \cite{iso_metric} evaluated at specific decision thresholds corresponding to fixed False Match Rates (FMR) \cite{iso_metric}. In accordance with established SOTA FIQA methodologies \cite{10449044, boutros_2023_crfiqa, MagFace}, we plot the EDC curves across all benchmarks at two fixed FMRs: $10^{-3}$ and $10^{-4}$.
Furthermore, we calculate both the Area Under the Curve (AUC) (in supplementary) and the partial AUC (pAUC) for the plotted EDC curves, providing a quantitative metric of verification performance across all rejection thresholds.
For better readability, we show $pAUC * 10^3$ and $AUC * 10^3$ values, which we will refer to as pAUC and AUC in the paper.
Following standard practices in the literature, the pAUC is evaluated up to a $30\%$ discard rate \cite{10449044, babnikTBIOM2024, DBLP:journals/tbbis/SchlettRTB24}.

\textbf{FR Models.}
To evaluate the generalizability of PreFIQs, we report verification performance across different quality discard rates using four distinct FR models: ArcFace \cite{deng2019arcface}, ElasticFace (ElasticFace-Arc) \cite{elasticface}, MagFace \cite{MagFace}, and CurricularFace \cite{curricularFace}.
For all experiments, we utilize the officially released pre-trained models provided by the respective authors \cite{curricularFace,elasticface,MagFace,deng2019arcface}.
Each model shares a ResNet100 backbone \cite{he_2016_resnet} originally trained on the MS1MV2 dataset \cite{guo_2016_ms1m, deng2019arcface}, and processes $112 \times 112$ aligned and cropped input images to generate 512-dimensional feature embeddings.

We evaluate these models under two distinct protocols: \textit{same-model} and \textit{cross-model}. Under the same-model protocol, ArcFace \cite{deng2019arcface} is employed both to compute the image quality scores and to execute the subsequent verification task. Under the cross-model protocol, ArcFace is used exclusively as the quality estimator to establish the discard rankings, while ElasticFace \cite{elasticface}, MagFace \cite{MagFace}, and CurricularFace \cite{curricularFace} act as the independent verification models evaluating the remaining image pairs.

\textbf{Comparisons with SOTA FIQA.}
We compare our PreFIQs approach against twelve SOTA FIQA methods: RankIQ \cite{RANKIQ_FIQA}, PFE \cite{PFE_FIQA}, SDD-FIQA \cite{SDDFIQA}, MagFace \cite{MagFace}, CR-FIQA \cite{boutros_2023_crfiqa}, DifFIQA \cite{10449044}, eDifFIQA \cite{babnikTBIOM2024}, CLIB-FIQA \cite{Ou_2024_CVPR}, VIT-FIQA \cite{atzori2025vitfiqaassessingfaceimage} as supervised approaches, SER-FIQ \cite{SERFIQ}, FaceQnet (v1 \cite{faceqnetv1}) \cite{hernandez2019faceqnet,faceqnetv1}, GraFIQs \cite{grafiqs}, ViTNT-FIQA \cite{vitnt_fiqa} as unsupervised approaches, and FROQ \cite{FROQ} as a semi-supervised approach.
A conceptual overview of PreFIQs and SOTA FIQA approaches is given in Table~\ref{tab:methods_info}.

\begin{table}[t]
  \centering
  \caption{Comparison overview of the operation and concepts of various FIQA approaches with our PreFIQs. Unsupervised approaches are labeled with \textcolor{blue!60}{BLUE}, and supervised and self-supervised methods are labeled with \textcolor{green!60}{GREEN} stripes, respectively.
  }
  %\vspace{-2mm}
  \resizebox{\columnwidth}{!}{%
    \begin{NiceTabular}{c l | *{4}{c} c c c c c}
         & \multicolumn{5}{c}{~} & \multicolumn{5}{c}{\textbf{Inference}} \\
             \cline{7-11}
        & \textbf{Method} 
        & \textbf{\rotatebox{90}{\makecell[l]{Quality\\Labels}}} 
        & \textbf{\rotatebox{90}{\makecell[l]{Architecture\\Specific}}} 
        & \textbf{\rotatebox{90}{\makecell[l]{Additional\\Training}}} 
        & \textbf{\rotatebox{90}{\makecell[l]{Custom\\Loss}}} 
        & \textbf{\rotatebox{90}{\makecell[l]{Feed-\\Forward}}} 
        & \textbf{\rotatebox{90}{\makecell[l]{Backwards}}}
        & \textbf{\rotatebox{90}{\makecell[l]{Feature\\Level}}}
        & \textbf{\rotatebox{90}{\makecell[l]{Gradient\\Level}}}
         & \textbf{\rotatebox{90}{\makecell[l]{Representation\\Level}}}\\
        \midrule
      
      \Block[tikz={pattern = {Lines[angle=-45, distance=1.0mm,  line width=0.5mm]},pattern color=green!40}]{7-1}{} 
      & SDD-FIQA~\cite{SDDFIQA}     & \ding{51} & \ding{55} & \ding{51} & \ding{55} & 1   & 0   & \ding{51} & \ding{55} & \ding{55} \\
      & PCNet~\cite{pcnet}           & \ding{51} & \ding{55} & \ding{51} & \ding{55} & 1   & 0   & \ding{51} & \ding{55} & \ding{55} \\
      & eDiFFIQA(L)~\cite{babnikTBIOM2024}& \ding{51} & \ding{55} & \ding{51} & \ding{51} & 1   & 0   & \ding{51} & \ding{55} & \ding{55} \\
      & CLIB-FIQA~\cite{Ou_2024_CVPR}   & \ding{51} & \ding{51} & \ding{51} & \ding{51} & 1   & 0   & \ding{51} & \ding{55} & \ding{55} \\
      \dashmidrule
      & MagFace~\cite{MagFace}       & \ding{55} & \ding{55} & \ding{51} & \ding{51} & 1   & 0   & \ding{51} & \ding{55} & \ding{55} \\
      & CR-FIQA~\cite{boutros_2023_crfiqa}       & \ding{55} & \ding{55} & \ding{51} & \ding{51} & 1   & 0   & \ding{51} & \ding{55} & \ding{55} \\
      & ViT-FIQA(T)~\cite{atzori2025vitfiqaassessingfaceimage} & \ding{55} & \ding{55} & \ding{51} & \ding{51} & 1   & 0   & \ding{51} & \ding{55} & \ding{55} \\
      \midrule
      \Block[tikz={preaction={fill, blue!40}, pattern = {Lines[angle=-45, distance=1.0mm,  line width=0.5mm]},pattern color=green!40}]{1-1}{} 
      & FROQ~\cite{FROQ}  & \ding{51} & \ding{55} & \ding{55} & \ding{55} & 1   & 0   & \ding{55} & \ding{55} & \ding{51} \\
      \midrule
      \Block[tikz={pattern = {Lines[angle=-45, distance=1.0mm,  line width=0.5mm]},pattern color=blue!40}]{5-1}{} 
      & SER-FIQ~\cite{SERFIQ}       & \ding{55} & \ding{51} & \ding{55} & \ding{55} & 100 & 0   & \ding{51} & \ding{55} & \ding{55} \\
      & FaceQAN~\cite{FaceQAN}      & \ding{55} & \ding{55} & \ding{55} & \ding{55} & 10  & 10  & \ding{51} & \ding{51} & \ding{55} \\
      & GraFIQs~\cite{grafiqs}      & \ding{55} & \ding{55} & \ding{55} & \ding{55} & 1   & 1   & \ding{55} & \ding{51} & \ding{55} \\
      & ViTNT-FIQA~\cite{vitnt_fiqa} & \ding{55} & \ding{55} & \ding{55} & \ding{55} & 1   & 0   & \ding{51} & \ding{55} & \ding{55} \\
      \midrule
      \rowcolor{gray!10}
      & PreFIQs (Ours) & \ding{55} & \ding{55} & \ding{55} & \ding{55} & 2   & 0   & \ding{51} & \ding{55} & \ding{55} \\
      \bottomrule
    \end{NiceTabular}%
  }%\vspace{-5mm}
  \label{tab:methods_info}
\end{table}

\section{Results}
\label{sec:results}

This section provides extensive overview of our results.
We first provide a quantitatively validation of the Jacobian-Vector product approximation introduced in Section~\ref{sec:methodology_subsubseec:theoretical_validation}, with results outlined in Table~\ref{tab:jacobian_validation}.
We then provide extensive overview of the FIQA performance when using different granularity of pruning (structured vs. unstructured), and different pruning criteria ($L_1$ magnitude vs. random pruning) across different pruning ratios $\rho$. The results of these experiments are shown in Table~\ref{tab:pruning_comparison_single_table_pauc_1e3}.
Additionally, we evaluate the impact of pruning on FR verification performance across a wide set of benchmarks, comparing the granularity of pruning and the pruning criteria across pruning ratios $\rho$.
The results are shown in Table~\ref{tab:fr_pruning_comparison}.
At the end of this Section, we compare our PreFIQs against recent SOTA approaches. The results for four FR models are shown in Table~\ref{tab:sota_pauc_1e3}.

\subsection{Jacobian-Vector Validation}
Table~\ref{tab:jacobian_validation} presents the quantitative comparison between the exact Jacobian-Vector product (Equation~\ref{eq:jacobian_drift}) and our proposed discrete representation drift $D(x)$ (Equation~\ref{eq:drift_calculation}).
The results show that both methods achieve nearly identical pAUC scores across all seven evaluation benchmarks and four FR models.
The average pAUC across all datasets and FR models is $10.514$ for the theoretical Jacobian drift and $10.558$ for our empirical discrete drift.
This strong alignment empirically validates our mathematical derivation.
It confirms that the computationally efficient PreFIQ, compared to the Jacobian-Vector product,  accurately approximates the geometric sensitivity of the latent manifold.
\begin{table}[!ht]
    \centering
\caption{Empirical validation of the Jacobian-Vector product approximation. At a sparsity ratio of $\rho=0.1$, the theoretical Jacobian-Vector product drift (Eq.~\ref{eq:jacobian_drift}) and the proposed discrete representation drift $\mathcal{D}(x)$ (Eq.~\ref{eq:drift_calculation}) achieve nearly identical pAUC scores (discard rate = $0.3$, across four used FR models). This quantitative alignment verifies that $\mathcal{D}(x)$ successfully captures the geometric sensitivity of the latent manifold.}
\label{tab:jacobian_validation}
%\vspace{-3mm}
    \resizebox{\columnwidth}{!}{%
\begin{NiceTabular}{c r |  r r r r r r r | r}
\Block{2-10}{\textbf{Average across FR Models} - $pAUC * 10^{3} \, ($FMR$=10^{-3}) \, [\downarrow]$} \\
 \\
 {} & \textbf{{Methods}} & \textbf{Adience} & \textbf{AgeDB-30} & \textbf{CFP-FP} & \textbf{LFW} & \textbf{CALFW} & \textbf{CPLFW} & \textbf{XQLFW} & $\overline{pAUC}$ \\
\midrule
\Block[tikz={pattern = {Lines[angle=-45, distance=1.5mm, line width=0.5mm]}, pattern color=gray!50}]{1-1} {} & \textbf{Jacobian Drift (Eq.~\ref{eq:jacobian_drift})} & $10.018$ & $6.863$ & $4.008$ & $0.858$ & $20.727$ & $20.608$ & $139.830$ & $\cellcolor{gray!50}10.514$ \\
\dashmidrule
\Block[tikz={pattern = {Dots[angle=45, distance=1.5mm, radius=0.3mm]}, pattern color=gray!50}]{1-1} {} & \textbf{Discrete Drift $D(x)$ (Eq.~\ref{eq:drift_calculation})} & $10.068$ & $6.913$ & $3.995$ & $0.858$ & $20.826$ & $20.689$ & $138.775$ & $\cellcolor{gray!50}10.558$ \\
\bottomrule
\end{NiceTabular}
}
\label{tab:mathematical_validation_jacobian}
%\vspace{-3mm}
\end{table}

\subsection{Evaluation of Pruning Approaches}
\label{sec:results_subsec:pruning_evaluation}
As outlined in our experimental setup, we compare the effects of different pruning strategies across various pruning ratios $\rho$. The average FIQA results for all FR models and benchmarks are presented in Table~\ref{tab:pruning_comparison_single_table_pauc_1e3}.
This evaluation is divided into two main analytical comparisons.

\textbf{Granularity of Pruning (unstructured vs. structured).} The results clearly show that unstructured pruning consistently achieves the best performance across all tested ratios. It reaches the best average pAUC of $10.516$ at a sparsity ratio of $\rho=0.4$. Furthermore, unstructured pruning maintains highly stable pAUC values across the majority of the tested spectrum. Performance degradation only becomes apparent at extremely high sparsity levels starting at $\rho=0.8$. In contrast, structured pruning achieves its best average pAUC of $11.137$ at the lowest sparsity setting of $\rho=0.1$ and shows significantly more sensitivity to increases in the pruning ratio. This steep performance decline can be attributed to the aggressive removal of entire architectural structures from the network. More importantly, this is also attributed to the fact that unstructured pruning maintains, to a large extent, FR verification accuracies compared to structured pruning, as shown in Table~\ref{tab:fr_pruning_comparison} and discussed in detail in Section~\ref{sec:results_subsec:eval_fr_performance}.

\textbf{Pruning criterion ($L_1$ magnitude vs. random pruning).} Parameter selection based on $L_1$ magnitude vastly outperforms random parameter selection, as shown in Table \ref{tab:pruning_comparison_single_table_pauc_1e3}. Random pruning yields a significantly worse pAUC of $18.298$ at $\rho=0.1$, in comparison to $L_1$ magnitude at the same pruning ratio. This can be attributed to the lower FR verification accuracies when the model is pruned using random pruning compared to $L_1$ magnitude criterion, as shown in Table~\ref{tab:fr_pruning_comparison} and discussed in detail in Section~\ref{sec:results_subsec:eval_fr_performance}.
Interestingly, after an initial performance drop, random pruning remains relatively stable across higher sparsity ratios compared to structured pruning. This suggests that pruning random parameters fails to isolate the critical network capacity responsible for encoding Pruning Identified Exemplars (Section~\ref{sec:methodology}), which ultimately results in a poor utility score.
\begin{table}[!ht]
    \centering
\caption{FIQ: unstructured vs. structured, and structured $L_1$ magnitude vs. structured random model pruning across different pruning ratios on four evaluated FR models reported using pAUC  (discard rate = 0.3, FMR = $10^{-3}$). The \textbf{best} and \textit{second-best} results per dataset are highlighted. The final column displays the average pAUC across all benchmarks. XQLFW is excluded from this average. Within this column, the best result is shaded per pruning category. 
It can be clearly observed that unstructured pruning led to better performance compared to structured pruning. In terms of pruning criterion,  $L_1$ magnitude outperformed, with a clear margin, random selection.  }
\resizebox{\columnwidth}{!}{%
\begin{NiceTabular}{c r |  r r r r r r r | r}

\Block{2-10}{\textbf{Granularity of Pruning - Unstructured vs. structured model pruning} $pAUC * 10^{3} \, ($FMR$=10^{-3}) \, [\downarrow]$} \\
 \\
 {} & \textbf{{Methods}} & \textbf{Adience} & \textbf{AgeDB-30} & \textbf{CFP-FP} & \textbf{LFW} & \textbf{CALFW} & \textbf{CPLFW} & \textbf{XQLFW} & $\overline{pAUC}$ \\
\midrule
\Block[tikz={pattern = {Lines[angle=-45, distance=1.5mm, line width=0.5mm]}, pattern color=cyan!50}]{9-1} {} & \textbf{Unstructured $\rho=$0.1} & $9.924$ & $\mathbf{6.809}$ & $3.846$ & $0.866$ & $21.435$ & $\mathit{20.442}$ & $137.081$ & $10.554$ \\
{} & \textbf{Unstructured $\rho=$0.2} & $10.123$ & $7.420$ & $\mathbf{3.719}$ & $0.895$ & $21.499$ & $\mathbf{20.099}$ & $\mathbf{134.395}$ & $10.626$ \\
{} & \textbf{Unstructured $\rho=$0.3} & $9.803$ & $6.961$ & $4.040$ & $\mathit{0.752}$ & $21.724$ & $20.473$ & $\mathit{136.922}$ & $10.626$ \\
{} & \textbf{Unstructured $\rho=$0.4} & $9.947$ & $6.982$ & $3.900$ & $0.899$ & $\mathbf{20.715}$ & $20.651$ & $139.022$ & $\cellcolor{cyan!10}10.516$ \\
{} & \textbf{Unstructured $\rho=$0.5} & $\mathbf{9.750}$ & $7.049$ & $\mathit{3.759}$ & $0.871$ & $21.081$ & $20.629$ & $139.902$ & $10.523$ \\
{} & \textbf{Unstructured $\rho=$0.6} & $\mathit{9.797}$ & $\mathit{6.867}$ & $4.027$ & $0.789$ & $\mathit{20.722}$ & $21.294$ & $144.214$ & $10.583$ \\
{} & \textbf{Unstructured $\rho=$0.7} & $10.060$ & $7.066$ & $4.579$ & $0.855$ & $21.336$ & $25.299$ & $150.031$ & $11.533$ \\
{} & \textbf{Unstructured $\rho=$0.8} & $11.684$ & $8.449$ & $8.290$ & $0.805$ & $21.841$ & $41.226$ & $161.248$ & $15.382$ \\
{} & \textbf{Unstructured $\rho=$0.9} & $14.758$ & $9.464$ & $11.722$ & $0.867$ & $22.995$ & $58.045$ & $177.987$ & $19.642$ \\
\dashmidrule
\Block[tikz={pattern = {Dots[angle=45, distance=1.5mm, radius=0.3mm]}, pattern color=orange!50}]{9-1} {} & \textbf{Structured $\rho=$0.1} & $10.695$ & $8.182$ & $4.452$ & $0.791$ & $21.137$ & $21.565$ & $144.276$ & $\cellcolor{orange!10}11.137$ \\
{} & \textbf{Structured $\rho=$0.2} & $12.066$ & $9.810$ & $9.218$ & $0.956$ & $21.925$ & $49.520$ & $157.267$ & $17.249$ \\
{} & \textbf{Structured $\rho=$0.3} & $16.232$ & $10.433$ & $12.477$ & $1.175$ & $23.157$ & $60.586$ & $169.934$ & $20.677$ \\
{} & \textbf{Structured $\rho=$0.4} & $16.651$ & $11.621$ & $12.596$ & $0.918$ & $24.447$ & $60.028$ & $170.537$ & $21.043$ \\
{} & \textbf{Structured $\rho=$0.5} & $16.515$ & $11.536$ & $12.023$ & $0.900$ & $24.002$ & $60.155$ & $171.338$ & $20.855$ \\
{} & \textbf{Structured $\rho=$0.6} & $16.189$ & $11.070$ & $11.985$ & $0.920$ & $24.049$ & $59.175$ & $171.926$ & $20.565$ \\
{} & \textbf{Structured $\rho=$0.7} & $17.402$ & $10.562$ & $12.064$ & $0.996$ & $24.250$ & $59.913$ & $170.943$ & $20.865$ \\
{} & \textbf{Structured $\rho=$0.8} & $16.504$ & $10.630$ & $12.155$ & $1.005$ & $24.195$ & $60.143$ & $172.842$ & $20.772$ \\
{} & \textbf{Structured $\rho=$0.9} & $16.801$ & $10.195$ & $12.556$ & $0.962$ & $24.011$ & $61.702$ & $176.857$ & $21.038$ \\
\bottomrule

\Block{2-10}{\textbf{Pruning Criterion - Comparison between $L_1$ magnitude and random model pruning} - $pAUC * 10^{3} \, ($FMR$=10^{-3}) \, [\downarrow]$} \\
 \\
 {} & \textbf{{Methods}} & \textbf{Adience} & \textbf{AgeDB-30} & \textbf{CFP-FP} & \textbf{LFW} & \textbf{CALFW} & \textbf{CPLFW} & \textbf{XQLFW} & $\overline{pAUC}$ \\
\midrule
 \Block[tikz={pattern = {Lines[angle=-45, distance=1.5mm, line width=0.5mm]}, pattern color=cyan!50}]{9-1} {} & \textbf{$L_1$ Magnitude\ $\rho=$0.1} & $9.924$ & $\mathbf{6.809}$ & $3.846$ & $0.866$ & $21.435$ & $\mathit{20.442}$ & $137.081$ & $10.554$ \\
{} & \textbf{$L_1$ Magnitude\ $\rho=$0.2} & $10.123$ & $7.420$ & $\mathbf{3.719}$ & $0.895$ & $21.499$ & $\mathbf{20.099}$ & $\mathbf{134.395}$ & $10.626$ \\
{} & \textbf{$L_1$ Magnitude\ $\rho=$0.3} & $9.803$ & $6.961$ & $4.040$ & $\mathit{0.752}$ & $21.724$ & $20.473$ & $\mathit{136.922}$ & $10.626$ \\
{} & \textbf{$L_1$ Magnitude\ $\rho=$0.4} & $9.947$ & $6.982$ & $3.900$ & $0.899$ & $\mathbf{20.715}$ & $20.651$ & $139.022$ & $\cellcolor{cyan!10}10.516$ \\
{} & \textbf{$L_1$ Magnitude\ $\rho=$0.5} & $\mathbf{9.750}$ & $7.049$ & $\mathit{3.759}$ & $0.871$ & $21.081$ & $20.629$ & $139.902$ & $10.523$ \\
{} & \textbf{$L_1$ Magnitude\ $\rho=$0.6} & $\mathit{9.797}$ & $\mathit{6.867}$ & $4.027$ & $0.789$ & $\mathit{20.722}$ & $21.294$ & $144.214$ & $10.583$ \\
{} & \textbf{$L_1$ Magnitude\ $\rho=$0.7} & $10.060$ & $7.066$ & $4.579$ & $0.855$ & $21.336$ & $25.299$ & $150.031$ & $11.533$ \\
{} & \textbf{$L_1$ Magnitude\ $\rho=$0.8} & $11.684$ & $8.449$ & $8.290$ & $0.805$ & $21.841$ & $41.226$ & $161.248$ & $15.382$ \\
{} & \textbf{$L_1$ Magnitude\ $\rho=$0.9} & $14.758$ & $9.464$ & $11.722$ & $0.867$ & $22.995$ & $58.045$ & $177.987$ & $19.642$ \\
\dashmidrule
\Block[tikz={pattern = {Hatch[angle=45, distance=1.5mm, line width=0.5mm]}, pattern color=violet!40}]{9-1} {} & \textbf{Random $\rho=$0.1} & $15.101$ & $9.258$ & $10.482$ & $\mathbf{0.729}$ & $22.696$ & $51.522$ & $181.241$ & $\cellcolor{violet!10}18.298$ \\
{} & \textbf{Random $\rho=$0.2} & $17.821$ & $10.157$ & $12.858$ & $1.135$ & $24.010$ & $61.720$ & $182.106$ & $21.283$ \\
{} & \textbf{Random $\rho=$0.3} & $16.707$ & $10.695$ & $12.351$ & $0.903$ & $23.764$ & $62.029$ & $176.676$ & $21.075$ \\
{} & \textbf{Random $\rho=$0.4} & $16.419$ & $10.792$ & $12.560$ & $0.883$ & $23.553$ & $61.756$ & $177.312$ & $20.994$ \\
{} & \textbf{Random $\rho=$0.5} & $15.986$ & $10.751$ & $12.271$ & $0.916$ & $23.650$ & $62.384$ & $179.484$ & $20.993$ \\
{} & \textbf{Random $\rho=$0.6} & $16.258$ & $11.102$ & $12.256$ & $0.921$ & $24.065$ & $61.378$ & $176.981$ & $20.997$ \\
{} & \textbf{Random $\rho=$0.7} & $16.568$ & $10.617$ & $12.211$ & $0.981$ & $23.791$ & $61.519$ & $178.097$ & $20.948$ \\
{} & \textbf{Random $\rho=$0.8} & $17.238$ & $10.988$ & $12.203$ & $0.900$ & $24.041$ & $61.293$ & $180.073$ & $21.110$ \\
{} & \textbf{Random $\rho=$0.9} & $17.221$ & $10.615$ & $12.025$ & $0.882$ & $24.163$ & $61.529$ & $180.387$ & $21.072$ \\
\bottomrule
\end{NiceTabular}
}
\label{tab:pruning_comparison_single_table_pauc_1e3}
%\vspace{-3mm}
\end{table}

\begin{table}[H]
    \centering
\caption{FIQ SOTA comparison using four FR models reported as pAUC scores (discard rate = 0.3, FMR = $10^{-3}$). The \textbf{best} and \textit{second-best} results per dataset are highlighted. The final column displays the average pAUC across all benchmarks. We exclude XQLFW from this average to prevent evaluation bias, as its quality labels were derived using SER-FIQ. The best average pAUC is highlighted in \begin{tabular}{c}\cellcolor{green!10}GREEN\end{tabular} for supervised and self-supervised approaches (marked using \textcolor{green!60}{green stripes}) , and \begin{tabular}{c}\cellcolor{blue!10}BLUE\end{tabular} for unsupervised approaches (marked with \textcolor{blue!60}{blue stripes}).
Our training-free PreFIQ is among the top-performing methods.
}
    \resizebox{\columnwidth}{!}{%
\begin{NiceTabular}{c r |  r r r r r r r r | r}
\Block{2-11}{\textbf{ArcFace~\cite{deng2019arcface}} - $pAUC * 10^{3} \, ($FMR$=10^{-3}) \, [\downarrow]$} \\
 \\
 {} & \textbf{{Methods}} & \textbf{Adience} & \textbf{AgeDB-30} & \textbf{CFP-FP} & \textbf{LFW} & \textbf{CALFW} & \textbf{CPLFW} & \textbf{XQLFW} & \textbf{IJB-C} & $\overline{pAUC}$ \\
\midrule
\Block[tikz={pattern = {Lines[angle=-45, distance=1.0mm,  line width=0.5mm]},pattern color=green!40}]{9-1} {} & \textbf{RankIQ~\cite{RANKIQ_FIQA}} & $14.572$ & $10.700$ & $8.323$ & $\mathit{0.748}$ & $22.522$ & $34.567$ & $148.080$ & $7.898$ & $14.190$ \\
{} & \textbf{PFE~\cite{PFE_FIQA}} & $10.740$ & $8.211$ & $5.893$ & $0.795$ & $22.256$ & $26.604$ & $142.459$ & $7.470$ & $11.710$ \\
{} & \textbf{SDD-FIQA~\cite{SDDFIQA}} & $11.844$ & $8.621$ & $7.810$ & $0.800$ & $22.354$ & $31.146$ & $159.151$ & $7.236$ & $12.830$ \\
{} & \textbf{MagFace~\cite{MagFace}} & $11.154$ & $7.428$ & $4.952$ & $\mathbf{0.680}$ & $21.066$ & $27.665$ & $160.833$ & $7.154$ & $11.443$ \\
{} & \textbf{CR-FIQA(L)~\cite{boutros_2023_crfiqa}} & $10.901$ & $7.605$ & $3.660$ & $0.810$ & $\mathit{20.937}$ & $20.374$ & $140.016$ & $6.579$ & $10.124$ \\
{} & \textbf{DifFIQA(R)~\cite{10449044}} & $11.226$ & $9.269$ & $3.931$ & $0.789$ & $21.801$ & $\mathit{20.185}$ & $137.822$ & $6.482$ & $10.526$ \\
{} & \textbf{eDifFIQA(L)~\cite{babnikTBIOM2024}} & $10.210$ & $\mathit{6.880}$ & $\mathbf{3.546}$ & $0.785$ & $21.012$ & $\mathbf{20.086}$ & $142.316$ & $\mathit{6.469}$ & $\cellcolor{green!10}9.856$ \\
{} & \textbf{CLIB-FIQA~\cite{Ou_2024_CVPR}} & $10.931$ & $7.387$ & $4.070$ & $0.790$ & $21.064$ & $20.431$ & $\mathit{137.399}$ & $6.596$ & $10.181$ \\
{} & \textbf{ViT-FIQA(T)~\cite{atzori2025vitfiqaassessingfaceimage}} & $\mathbf{9.948}$ & $8.234$ & $\mathit{3.568}$ & $0.771$ & $21.771$ & $20.531$ & $140.465$ & $6.563$ & $10.198$ \\
\dashmidrule
\Block[tikz={preaction={fill, blue!40}, pattern = {Lines[angle=-45, distance=1.0mm,  line width=0.5mm]},pattern color=green!40}]{1-1} {} & \textbf{FROQ~\cite{FROQ}} & $12.463$ & $8.297$ & $5.550$ & $0.835$ & $\mathbf{20.914}$ & $22.968$ & $140.843$ & $\mathbf{6.438}$ & $11.066$ \\
\dashmidrule
\Block[tikz={pattern = {Lines[angle=-45, distance=1.0mm,  line width=0.5mm]},pattern color=blue!40}]{4-1} {} & \textbf{SER-FIQ~\cite{SERFIQ}} & $11.627$ & $7.776$ & $3.797$ & $0.800$ & $22.053$ & $21.570$ & $\mathbf{132.368}$ & $6.528$ & $\cellcolor{blue!10}10.593$ \\
{} & \textbf{FaceQnet~\cite{hernandez2019faceqnet,faceqnetv1}} & $15.273$ & $8.804$ & $9.009$ & $1.007$ & $23.339$ & $50.881$ & $183.144$ & $8.502$ & $16.688$ \\
{} & \textbf{GraFIQs(L)~\cite{grafiqs}} & $10.541$ & $7.717$ & $4.348$ & $0.840$ & $21.425$ & $22.495$ & $144.309$ & $6.863$ & $10.604$ \\
{} & \textbf{ViTNT-FIQA~\cite{vitnt_fiqa}} & $10.706$ & $9.674$ & $4.568$ & $0.988$ & $22.292$ & $21.802$ & $140.730$ & $6.732$ & $10.966$ \\
\dashmidrule
\rowcolor{gray!10}
\Block[tikz={pattern = {Lines[angle=-45, distance=1.0mm,  line width=0.5mm]},pattern color=blue!40}]{1-1} {} & \textbf{PreFIQs (Ours)} & $\mathit{10.009}$ & $\mathbf{6.876}$ & $3.755$ & $0.921$ & $20.979$ & $21.180$ & $141.716$ & $6.770$ & $\underline{10.070}$ \\
\bottomrule

\Block{2-11}{\textbf{CurricularFace~\cite{curricularFace}} - $pAUC * 10^{3} \, ($FMR$=10^{-3}) \, [\downarrow]$} \\
 \\
 {} & \textbf{{Methods}} & \textbf{Adience} & \textbf{AgeDB-30} & \textbf{CFP-FP} & \textbf{LFW} & \textbf{CALFW} & \textbf{CPLFW} & \textbf{XQLFW} & \textbf{IJB-C} & $\overline{pAUC}$ \\
\midrule
\Block[tikz={pattern = {Lines[angle=-45, distance=1.0mm,  line width=0.5mm]},pattern color=green!40}]{9-1} {} & \textbf{RankIQ~\cite{RANKIQ_FIQA}} & $12.521$ & $11.441$ & $9.088$ & $\mathit{0.748}$ & $21.544$ & $31.151$ & $132.029$ & $7.654$ & $13.449$ \\
{} & \textbf{PFE~\cite{PFE_FIQA}} & $9.523$ & $8.478$ & $6.497$ & $0.795$ & $21.723$ & $22.546$ & $120.134$ & $7.090$ & $10.950$ \\
{} & \textbf{SDD-FIQA~\cite{SDDFIQA}} & $10.522$ & $9.492$ & $8.394$ & $0.800$ & $21.750$ & $26.271$ & $142.492$ & $6.904$ & $12.019$ \\
{} & \textbf{MagFace~\cite{MagFace}} & $10.179$ & $7.652$ & $5.352$ & $\mathbf{0.680}$ & $20.796$ & $24.451$ & $150.727$ & $6.794$ & $10.844$ \\
{} & \textbf{CR-FIQA(L)~\cite{boutros_2023_crfiqa}} & $10.111$ & $7.556$ & $4.054$ & $0.830$ & $20.701$ & $17.364$ & $\mathit{119.232}$ & $6.305$ & $9.560$ \\
{} & \textbf{DifFIQA(R)~\cite{10449044}} & $9.806$ & $9.949$ & $3.772$ & $0.789$ & $21.065$ & $\mathit{17.046}$ & $124.298$ & $6.185$ & $9.802$ \\
{} & \textbf{eDifFIQA(L)~\cite{babnikTBIOM2024}} & $8.996$ & $7.576$ & $\mathbf{3.520}$ & $0.785$ & $20.597$ & $\mathbf{16.947}$ & $131.594$ & $\mathbf{6.164}$ & $\cellcolor{green!10}9.226$ \\
{} & \textbf{CLIB-FIQA~\cite{Ou_2024_CVPR}} & $9.768$ & $8.103$ & $3.841$ & $0.790$ & $\mathit{20.489}$ & $17.367$ & $123.186$ & $6.321$ & $9.526$ \\
{} & \textbf{ViT-FIQA(T)~\cite{atzori2025vitfiqaassessingfaceimage}} & $\mathbf{8.899}$ & $8.606$ & $3.973$ & $0.771$ & $21.439$ & $17.304$ & $124.911$ & $6.337$ & $9.618$ \\
\dashmidrule
\Block[tikz={preaction={fill, blue!40}, pattern = {Lines[angle=-45, distance=1.0mm,  line width=0.5mm]},pattern color=green!40}]{1-1} {} & \textbf{FROQ~\cite{FROQ}} & $10.656$ & $8.972$ & $7.177$ & $0.835$ & $\mathbf{20.212}$ & $19.619$ & $125.059$ & $\mathit{6.174}$ & $10.521$ \\
\dashmidrule
\Block[tikz={pattern = {Lines[angle=-45, distance=1.0mm,  line width=0.5mm]},pattern color=blue!40}]{4-1} {} & \textbf{SER-FIQ~\cite{SERFIQ}} & $10.404$ & $7.796$ & $3.811$ & $0.851$ & $21.193$ & $18.447$ & $\mathbf{117.754}$ & $6.253$ & $\cellcolor{blue!10}9.822$ \\
{} & \textbf{FaceQnet~\cite{hernandez2019faceqnet,faceqnetv1}} & $13.608$ & $9.627$ & $8.580$ & $1.007$ & $22.762$ & $42.412$ & $159.222$ & $8.081$ & $15.154$ \\
{} & \textbf{GraFIQs(L)~\cite{grafiqs}} & $9.694$ & $\mathit{7.449}$ & $4.081$ & $0.880$ & $20.886$ & $19.500$ & $125.193$ & $6.494$ & $9.855$ \\
{} & \textbf{ViTNT-FIQA~\cite{vitnt_fiqa}} & $9.625$ & $10.058$ & $5.358$ & $0.988$ & $21.549$ & $18.389$ & $129.235$ & $6.383$ & $10.336$ \\
\dashmidrule
\rowcolor{gray!10}
\Block[tikz={pattern = {Lines[angle=-45, distance=1.0mm,  line width=0.5mm]},pattern color=blue!40}]{1-1} {} & \textbf{PreFIQs (Ours)} & $\mathit{8.968}$ & $\mathbf{7.020}$ & $\mathit{3.752}$ & $0.921$ & $20.577$ & $18.239$ & $123.709$ & $6.445$ & $\underline{9.417}$ \\
\bottomrule

\Block{2-11}{\textbf{ElasticFace~\cite{elasticface}} - $pAUC * 10^{3} \, ($FMR$=10^{-3}) \, [\downarrow]$} \\
 \\
 {} & \textbf{{Methods}} & \textbf{Adience} & \textbf{AgeDB-30} & \textbf{CFP-FP} & \textbf{LFW} & \textbf{CALFW} & \textbf{CPLFW} & \textbf{XQLFW} & \textbf{IJB-C} & $\overline{pAUC}$ \\
\midrule
\Block[tikz={pattern = {Lines[angle=-45, distance=1.0mm,  line width=0.5mm]},pattern color=green!40}]{9-1} {} & \textbf{RankIQ~\cite{RANKIQ_FIQA}} & $16.208$ & $10.588$ & $7.716$ & $\mathit{0.578}$ & $21.469$ & $32.959$ & $134.681$ & $7.737$ & $13.894$ \\
{} & \textbf{PFE~\cite{PFE_FIQA}} & $11.804$ & $7.437$ & $5.414$ & $0.678$ & $21.482$ & $23.735$ & $133.068$ & $7.062$ & $11.088$ \\
{} & \textbf{SDD-FIQA~\cite{SDDFIQA}} & $13.253$ & $8.766$ & $6.139$ & $0.681$ & $21.410$ & $28.248$ & $157.945$ & $6.993$ & $12.213$ \\
{} & \textbf{MagFace~\cite{MagFace}} & $12.355$ & $6.954$ & $4.767$ & $\mathbf{0.564}$ & $20.546$ & $26.468$ & $158.092$ & $6.907$ & $11.223$ \\
{} & \textbf{CR-FIQA(L)~\cite{boutros_2023_crfiqa}} & $11.770$ & $7.367$ & $3.288$ & $0.692$ & $\mathit{20.193}$ & $19.265$ & $\mathit{124.870}$ & $6.355$ & $9.847$ \\
{} & \textbf{DifFIQA(R)~\cite{10449044}} & $12.572$ & $8.553$ & $3.460$ & $0.685$ & $20.990$ & $\mathit{18.780}$ & $127.943$ & $\mathit{6.240}$ & $10.183$ \\
{} & \textbf{eDifFIQA(L)~\cite{babnikTBIOM2024}} & $\mathit{11.193}$ & $\mathbf{6.587}$ & $\mathbf{3.040}$ & $0.681$ & $20.246$ & $\mathbf{18.774}$ & $135.841$ & $\mathbf{6.197}$ & $\cellcolor{green!10}9.531$ \\
{} & \textbf{CLIB-FIQA~\cite{Ou_2024_CVPR}} & $11.808$ & $7.144$ & $3.411$ & $0.674$ & $20.196$ & $19.231$ & $129.072$ & $6.397$ & $9.837$ \\
{} & \textbf{ViT-FIQA(T)~\cite{atzori2025vitfiqaassessingfaceimage}} & $11.228$ & $7.607$ & $\mathit{3.200}$ & $0.654$ & $20.764$ & $19.469$ & $135.159$ & $6.334$ & $9.894$ \\
\dashmidrule
\Block[tikz={preaction={fill, blue!40}, pattern = {Lines[angle=-45, distance=1.0mm,  line width=0.5mm]},pattern color=green!40}]{1-1} {} & \textbf{FROQ~\cite{FROQ}} & $13.919$ & $8.420$ & $5.166$ & $0.718$ & $\mathbf{20.027}$ & $21.892$ & $136.545$ & $6.242$ & $10.912$ \\
\dashmidrule
\Block[tikz={pattern = {Lines[angle=-45, distance=1.0mm,  line width=0.5mm]},pattern color=blue!40}]{4-1} {} & \textbf{SER-FIQ~\cite{SERFIQ}} & $12.933$ & $7.430$ & $3.428$ & $0.735$ & $20.911$ & $20.168$ & $\mathbf{118.090}$ & $6.328$ & $10.276$ \\
{} & \textbf{FaceQnet~\cite{hernandez2019faceqnet,faceqnetv1}} & $16.806$ & $8.696$ & $8.261$ & $0.890$ & $22.592$ & $44.257$ & $171.840$ & $8.250$ & $15.679$ \\
{} & \textbf{GraFIQs(L)~\cite{grafiqs}} & $11.348$ & $7.678$ & $3.757$ & $0.724$ & $20.796$ & $21.031$ & $146.097$ & $6.536$ & $\cellcolor{blue!10}10.267$ \\
{} & \textbf{ViTNT-FIQA~\cite{vitnt_fiqa}} & $12.031$ & $9.141$ & $4.136$ & $0.832$ & $21.377$ & $20.535$ & $137.532$ & $6.431$ & $10.640$ \\
\dashmidrule
\rowcolor{gray!10}
\Block[tikz={pattern = {Lines[angle=-45, distance=1.0mm,  line width=0.5mm]},pattern color=blue!40}]{1-1} {} & \textbf{PreFIQs (Ours)} & $\mathbf{10.664}$ & $\mathit{6.600}$ & $3.287$ & $0.805$ & $20.324$ & $19.926$ & $135.726$ & $6.461$ & $\underline{9.724}$ \\
\bottomrule

\Block{2-11}{\textbf{MagFace~\cite{MagFace}} - $pAUC * 10^{3} \, ($FMR$=10^{-3}) \, [\downarrow]$} \\
 \\
 {} & \textbf{{Methods}} & \textbf{Adience} & \textbf{AgeDB-30} & \textbf{CFP-FP} & \textbf{LFW} & \textbf{CALFW} & \textbf{CPLFW} & \textbf{XQLFW} & \textbf{IJB-C} & $\overline{pAUC}$ \\
\midrule
\Block[tikz={pattern = {Lines[angle=-45, distance=1.0mm,  line width=0.5mm]},pattern color=green!40}]{9-1} {} & \textbf{RankIQ~\cite{RANKIQ_FIQA}} & $14.574$ & $11.894$ & $11.164$ & $0.812$ & $22.222$ & $37.772$ & $162.060$ & $9.217$ & $15.379$ \\
{} & \textbf{PFE~\cite{PFE_FIQA}} & $10.996$ & $8.598$ & $7.291$ & $0.804$ & $22.160$ & $27.130$ & $158.349$ & $8.461$ & $12.206$ \\
{} & \textbf{SDD-FIQA~\cite{SDDFIQA}} & $12.131$ & $9.542$ & $9.532$ & $0.826$ & $22.149$ & $31.445$ & $180.623$ & $8.414$ & $13.434$ \\
{} & \textbf{MagFace~\cite{MagFace}} & $11.258$ & $\mathit{7.504}$ & $6.264$ & $\mathbf{0.706}$ & $21.071$ & $29.659$ & $176.288$ & $8.223$ & $12.098$ \\
{} & \textbf{CR-FIQA(L)~\cite{boutros_2023_crfiqa}} & $11.321$ & $8.140$ & $4.993$ & $0.818$ & $21.029$ & $22.242$ & $151.180$ & $7.759$ & $10.900$ \\
{} & \textbf{DifFIQA(R)~\cite{10449044}} & $11.482$ & $9.818$ & $5.447$ & $0.815$ & $21.687$ & $22.351$ & $151.131$ & $7.603$ & $11.315$ \\
{} & \textbf{eDifFIQA(L)~\cite{babnikTBIOM2024}} & $10.614$ & $7.699$ & $\mathbf{4.756}$ & $0.810$ & $21.085$ & $\mathbf{22.186}$ & $161.214$ & $\mathbf{7.554}$ & $\cellcolor{green!10}10.672$ \\
{} & \textbf{CLIB-FIQA~\cite{Ou_2024_CVPR}} & $11.301$ & $8.128$ & $5.317$ & $0.799$ & $\mathit{20.967}$ & $22.665$ & $\mathit{149.878}$ & $7.715$ & $10.985$ \\
{} & \textbf{ViT-FIQA(T)~\cite{atzori2025vitfiqaassessingfaceimage}} & $\mathit{10.184}$ & $8.644$ & $4.926$ & $\mathit{0.779}$ & $21.611$ & $\mathit{22.228}$ & $151.108$ & $7.686$ & $10.865$ \\
\dashmidrule
\Block[tikz={preaction={fill, blue!40}, pattern = {Lines[angle=-45, distance=1.0mm,  line width=0.5mm]},pattern color=green!40}]{1-1} {} & \textbf{FROQ~\cite{FROQ}} & $12.556$ & $9.524$ & $8.486$ & $0.843$ & $\mathbf{20.786}$ & $24.584$ & $159.701$ & $\mathit{7.579}$ & $12.051$ \\
\dashmidrule
\Block[tikz={pattern = {Lines[angle=-45, distance=1.0mm,  line width=0.5mm]},pattern color=blue!40}]{4-1} {} & \textbf{SER-FIQ~\cite{SERFIQ}} & $12.104$ & $8.696$ & $4.918$ & $0.877$ & $21.650$ & $23.579$ & $\mathbf{144.182}$ & $7.660$ & $11.355$ \\
{} & \textbf{FaceQnet~\cite{hernandez2019faceqnet,faceqnetv1}} & $15.601$ & $9.546$ & $11.081$ & $1.015$ & $22.995$ & $75.482$ & $190.321$ & $9.644$ & $20.766$ \\
{} & \textbf{GraFIQs(L)~\cite{grafiqs}} & $10.985$ & $8.041$ & $5.454$ & $0.921$ & $21.267$ & $24.745$ & $160.852$ & $8.024$ & $\cellcolor{blue!10}11.348$ \\
{} & \textbf{ViTNT-FIQA~\cite{vitnt_fiqa}} & $11.043$ & $9.897$ & $6.528$ & $0.974$ & $22.024$ & $23.590$ & $151.360$ & $7.782$ & $11.691$ \\
\dashmidrule
\rowcolor{gray!10}
\Block[tikz={pattern = {Lines[angle=-45, distance=1.0mm,  line width=0.5mm]},pattern color=blue!40}]{1-1} {} & \textbf{PreFIQs (Ours)} & $\mathbf{10.146}$ & $\mathbf{7.432}$ & $\mathit{4.805}$ & $0.947$ & $20.981$ & $23.261$ & $154.940$ & $7.838$ & $\underline{10.773}$ \\
\bottomrule
\end{NiceTabular}
}
\label{tab:sota_pauc_1e3}
%\vspace{-3mm}
\end{table}

\subsection{Evaluation of FR Performance}
\label{sec:results_subsec:eval_fr_performance}
Table~\ref{tab:fr_pruning_comparison} compares the underlying FR verification accuracy of the used FR models across different granularities of pruning and pruning criteria across different pruning ratios. 

\textbf{Granularity of Pruning:} The results demonstrate that unstructured pruning significantly outperforms structured pruning. Unstructured pruning maintains a highly consistent verification performance across most pruning ratios, experiencing a notable drop only at the extreme ratio of $\rho=0.9$. In contrast, structured pruning suffers a severe and rapid degradation in accuracy as entire architectural components are removed from the network.

\textbf{Pruning criterion:} Pruning parameters based on $L_1$ magnitude drastically outperforms random pruning. Unstructured random pruning begins to lose its discriminative power almost immediately and completely collapses to random guessing (accuracy $50.00\%$) at a relatively low sparsity ratio of $\rho=0.5$. 
This rapid decline in the accuracy of FR verification is directly correlated and explains the corresponding loss in the FIQA performance observed for structured and random pruning strategies discussed in the previous Section~\ref{sec:results_subsec:pruning_evaluation}.
\begin{table}[!ht]
    \centering
    \caption{FR verification accuracy (\%) of ResNet100 under different pruning strategies at pruning ratios $\rho$. The \textbf{best} and \textit{second-best} results per dataset are highlighted. %
    It is evident that unstructured pruning consistently achieves superior performance compared to structured pruning. Regarding the pruning criterion, $L_1$-magnitude-based pruning clearly outperforms random parameter selection by a substantial margin.}
    \resizebox{\columnwidth}{!}{%
    \begin{NiceTabular}{c r |  r r r r r r | r}
    \Block{2-9}{\textbf{Granularity of Pruning - Comparison between unstructured and structured model pruning} $[\uparrow]$} \\
    \\
    {} & \textbf{Methods} & \textbf{LFW} & \textbf{CFP-FP} & \textbf{CFP-FF} & \textbf{AgeDB-30} & \textbf{CALFW} & \textbf{CPLFW} & $\overline{\text{Acc}}$ [$\uparrow$] \\
    \midrule
    \Block[tikz={pattern = {Lines[angle=-45, distance=1.5mm, line width=0.5mm]}, pattern color=gray!50}]{1-1} {} & \textbf{ResNet100 (unpruned)} & $\mathbf{99.80}$ & $\mathit{96.67}$ & $\mathit{99.89}$ & $98.35$ & $\mathit{96.15}$ & $\mathbf{93.32}$ & $\cellcolor{gray!30}97.36$ \\
    \dashmidrule
    \Block[tikz={pattern = {Lines[angle=-45, distance=1.5mm, line width=0.5mm]}, pattern color=cyan!50}]{5-1} {} & \textbf{Unstructured\ $\rho{=}0.1$} & $\mathbf{99.80}$ & $\mathit{96.67}$ & $\mathit{99.89}$ & $\mathbf{98.43}$ & $\mathbf{96.17}$ & $\mathit{93.23}$ &  $\cellcolor{cyan!10}97.37$ \\
    {} & \textbf{Unstructured\ $\rho{=}0.3$} & $\mathbf{99.80}$ & $96.59$ & $\mathit{99.89}$ & $\mathit{98.42}$ & $96.08$ & $93.17$ & $97.32$ \\
    {} & \textbf{Unstructured\ $\rho{=}0.5$} & $\mathbf{99.80}$ & $96.36$ & $\mathbf{99.90}$ & $98.20$ & $96.00$ & $92.70$ & $97.16$ \\
    {} & \textbf{Unstructured\ $\rho{=}0.7$} & $99.75$ & $94.87$ & $99.79$ & $97.50$ & $95.85$ & $90.38$ & $96.36$ \\
    {} & \textbf{Unstructured\ $\rho{=}0.9$} & $90.45$ & $65.76$ & $91.87$ & $75.27$ & $78.37$ & $58.87$ & $76.76$ \\
    \dashmidrule
    \Block[tikz={pattern = {Dots[angle=45, distance=1.5mm, radius=0.3mm]}, pattern color=orange!50}]{5-1} {} & \textbf{Structured\ $\rho{=}0.1$} & $\mathit{99.77}$ & $95.30$ & $99.84$ & $97.82$ & $95.88$ & $91.07$ & $\cellcolor{orange!10}96.61$ \\
    {} & \textbf{Structured\ $\rho{=}0.3$} & $82.90$ & $58.71$ & $79.10$ & $71.93$ & $68.22$ & $56.72$ & $69.60$ \\
    {} & \textbf{Structured\ $\rho{=}0.5$} & $72.47$ & $59.39$ & $71.50$ & $58.35$ & $58.92$ & $52.77$ & $62.23$ \\
    {} & \textbf{Structured\ $\rho{=}0.7$} & $51.12$ & $50.19$ & $51.26$ & $50.10$ & $50.17$ & $50.60$ & $50.57$ \\
    {} & \textbf{Structured\ $\rho{=}0.9$} & $50.00$ & $50.00$ & $50.00$ & $50.00$ & $50.00$ & $50.00$ & $50.00$ \\
    \bottomrule

    \Block{2-9}{\textbf{Pruning Criterion - Comparison between $L_1$ magnitude and random model pruning} $[\uparrow]$} \\
    \\
    {} & \textbf{Methods} & \textbf{LFW} & \textbf{CFP-FP} & \textbf{CFP-FF} & \textbf{AgeDB-30} & \textbf{CALFW} & \textbf{CPLFW} & $\overline{\text{Acc}}$ [$\uparrow$] \\
    \midrule
    \Block[tikz={pattern = {Lines[angle=-45, distance=1.5mm, line width=0.5mm]}, pattern color=gray!50}]{1-1} {} & \textbf{ResNet100 (unpruned)} & $\mathbf{99.80}$ & $\mathit{96.67}$ & $\mathit{99.89}$ & $98.35$ & $\mathit{96.15}$ & $\mathbf{93.32}$ & $\cellcolor{gray!30}97.36$ \\
    \dashmidrule
    \Block[tikz={pattern = {Lines[angle=-45, distance=1.5mm, line width=0.5mm]}, pattern color=cyan!50}]{5-1} {} & \textbf{$L_1$ Magnitude\ $\rho{=}0.1$} & $\mathbf{99.80}$ & $\mathit{96.67}$ & $\mathit{99.89}$ & $\mathbf{98.43}$ & $\mathbf{96.17}$ & $\mathit{93.23}$ & $\cellcolor{cyan!10}97.37$ \\
    {} & \textbf{$L_1$ Magnitude\ $\rho{=}0.3$} & $\mathbf{99.80}$ & $96.59$ & $\mathit{99.89}$ & $\mathit{98.42}$ & $96.08$ & $93.17$ & $97.32$ \\
    {} & \textbf{$L_1$ Magnitude\ $\rho{=}0.5$} & $\mathbf{99.80}$ & $96.36$ & $\mathbf{99.90}$ & $98.20$ & $96.00$ & $92.70$ & $97.16$ \\
    {} & \textbf{$L_1$ Magnitude\ $\rho{=}0.7$} & $99.75$ & $94.87$ & $99.79$ & $97.50$ & $95.85$ & $90.38$ & $96.36$ \\
    {} & \textbf{$L_1$ Magnitude\ $\rho{=}0.9$} & $90.45$ & $65.76$ & $91.87$ & $75.27$ & $78.37$ & $58.87$ & $76.76$ \\
    \dashmidrule
    \Block[tikz={pattern = {Hatch[angle=45, distance=1.5mm, line width=0.5mm]}, pattern color=violet!40}]{5-1} {} & \textbf{Random $\rho{=}0.1$} & $97.43$ & $77.86$ & $98.03$ & $82.82$ & $89.18$ & $73.52$ & $\cellcolor{violet!10}86.47$ \\
    {} & \textbf{Random $\rho{=}0.3$} & $71.73$ & $59.83$ & $73.21$ & $56.63$ & $57.88$ & $55.07$ & $62.39$ \\
    {} & \textbf{Random $\rho{=}0.5$} & $50.00$ & $50.00$ & $50.00$ & $50.00$ & $50.00$ & $50.00$ & $50.00$ \\
    {} & \textbf{Random $\rho{=}0.7$} & $50.00$ & $50.00$ & $50.00$ & $50.00$ & $50.00$ & $50.00$ & $50.00$ \\
    {} & \textbf{Random $\rho{=}0.9$} & $50.00$ & $50.00$ & $50.00$ & $50.00$ & $50.00$ & $50.00$ & $50.00$ \\
    \bottomrule
    \end{NiceTabular}
    }
    \label{tab:fr_pruning_comparison}
\end{table}

\subsection{Comparison to State-of-the-Art}
Table~\ref{tab:sota_pauc_1e3} presents a comparison of our PreFIQs (unstructured $L_1$ magnitude pruning at $\rho=0.4$, our best setups Table \ref{tab:pruning_comparison_single_table_pauc_1e3}) against recent FIQA approaches across four  FR models.

The results demonstrate that PreFIQs achieves highly competitive performance compared to the top-performing SOTA methods. Most notably, PreFIQs establishes the new SOTA performance on the challenging AgeDB-30 benchmark across three evaluated FR models (ArcFace, CurricularFace, and MagFace), while achieving second-best performance on the ElasticFace model.
Furthermore, PreFIQs consistently achieves the top or second-best performance on the Adience across all four evaluated FR models.

Overall, our entirely training-free PreFIQs approach successfully outperforms several complex supervised approaches across multiple benchmarks. On the large-scale IJB-C dataset, PreFIQs yields highly competitive results, further validating the robustness and generalizability of parameter sparsification as a reliable metric for FIQA.

\begin{figure}[h!]
\caption{ EDC (FNMR at FMR=$1e^{-3}$) of PreFIQs and recent FIQA approaches.
    The results are shown for four FR models on eight benchmarks.
    Unsupervised approaches are visualized using dotted lines. Supervised methods are visualized with dashed lines.
    PreFIQs is visualized using a continuous line with shaded AUC.
    }
    \label{fig:edc_curves}%\vspace{-2mm}
    \centering
    \includegraphics[width=1.0\columnwidth]{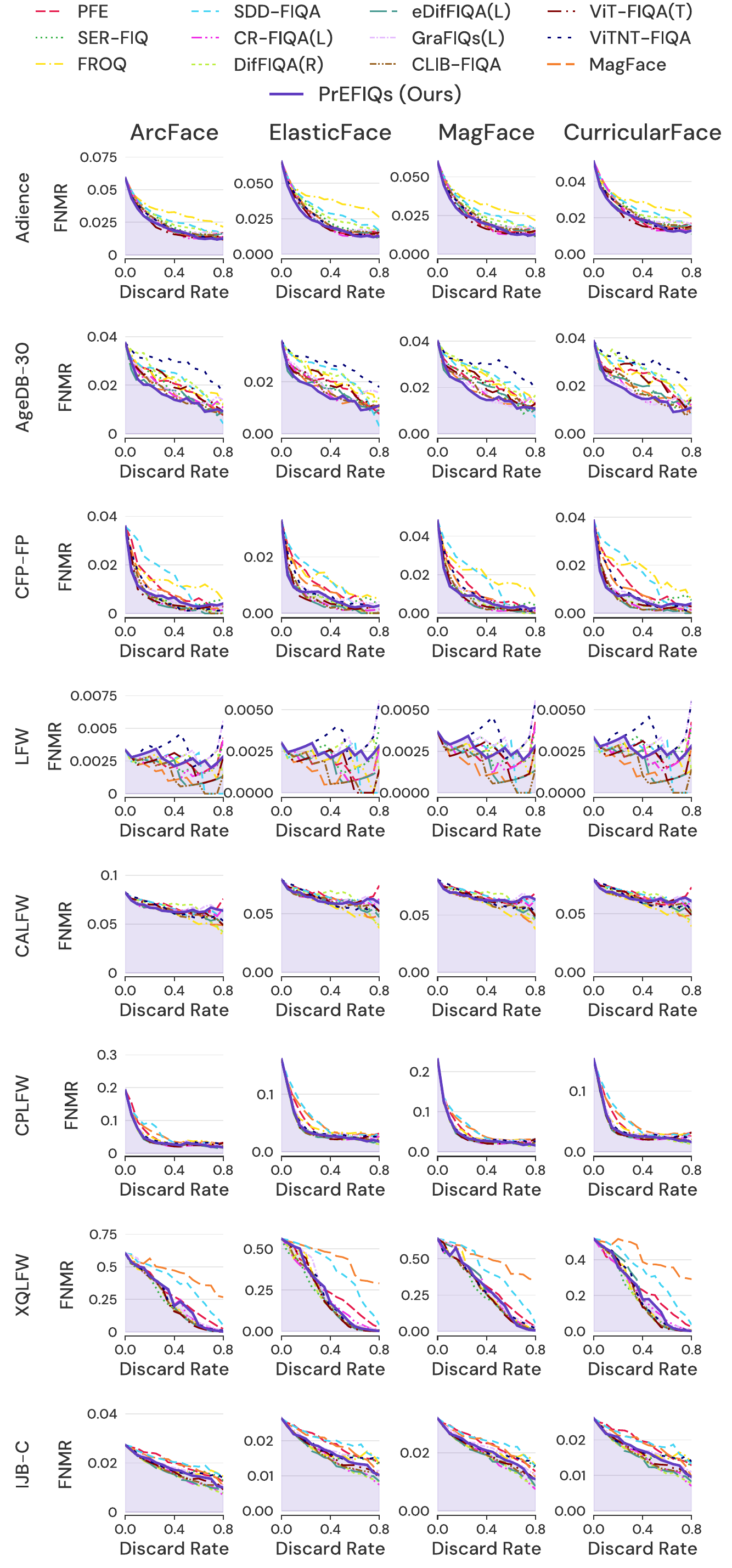}\vspace{-3mm}
\end{figure}

\vspace{-1mm}
\section{Conclusion}
\label{sec:conclusion}
This paper introduced PreFIQs, a novel data-free and training-free framework for FIQA. Departing from prior approaches, PreFIQs reframes image utility as structural robustness under model sparsification. Grounded in the PIE hypothesis, we demonstrated that the representation drift between an original FR model and its pruned counterpart provides a principled and computationally efficient proxy for image quality.
We provided both theoretical and empirical validation of this formulation. A first-order Taylor analysis showed that the proposed discrete embedding drift approximates the Jacobian-vector product governing geometric sensitivity of the latent identity manifold. Extensive experiments across eight benchmarks and four SOTA FR models confirmed this alignment, demonstrating that PreFIQs achieves highly competitive, and in several cases SOTA, performance, particularly on challenging benchmarks such as AgeDB-30 and Adience. 
Beyond its empirical effectiveness, PreFIQs offers a conceptual shift in FIQA: rather than predicting quality through learned regression or stochastic robustness estimation, it directly measures how well identity information survives controlled capacity reduction. This perspective establishes parameter sparsification as a probe of sample utility. Ultimately, our results support a simple but powerful principle: face image quality is what survives pruning.

\section*{Acknowledgment}
This research work has been funded by the German Federal Ministry of Education and Research and the Hessen State Ministry for Higher Education, Research and the Arts within their joint support of the National Research Center for Applied Cybersecurity ATHENE.
{
    \small
    \bibliographystyle{ieeenat_fullname}
    \bibliography{main}
}
\clearpage
\section{Supplementary Material}
\label{sec:supplementary_material_introduction}

\begin{figure*}
\caption{Comparison of EDC curves (FNMR at FMR=$1e^{-3}$) of PreFIQs against recent FIQA approaches.
    The results are shown for four FR models on eight benchmarks.
    Unsupervised approaches are visualized using dotted lines. Supervised methods are visualized with dashed lines.
    PreFIQs is visualized using a continuous line with shaded AUC.
    For PreFIQs, unstructured $L_1$ magnitude pruning with $\rho=0.4$ is used.
    }
    \label{fig:edc_curves_supp_fnmr3}\vspace{-2mm}
    \centering
    \includegraphics[width=1.0\linewidth]{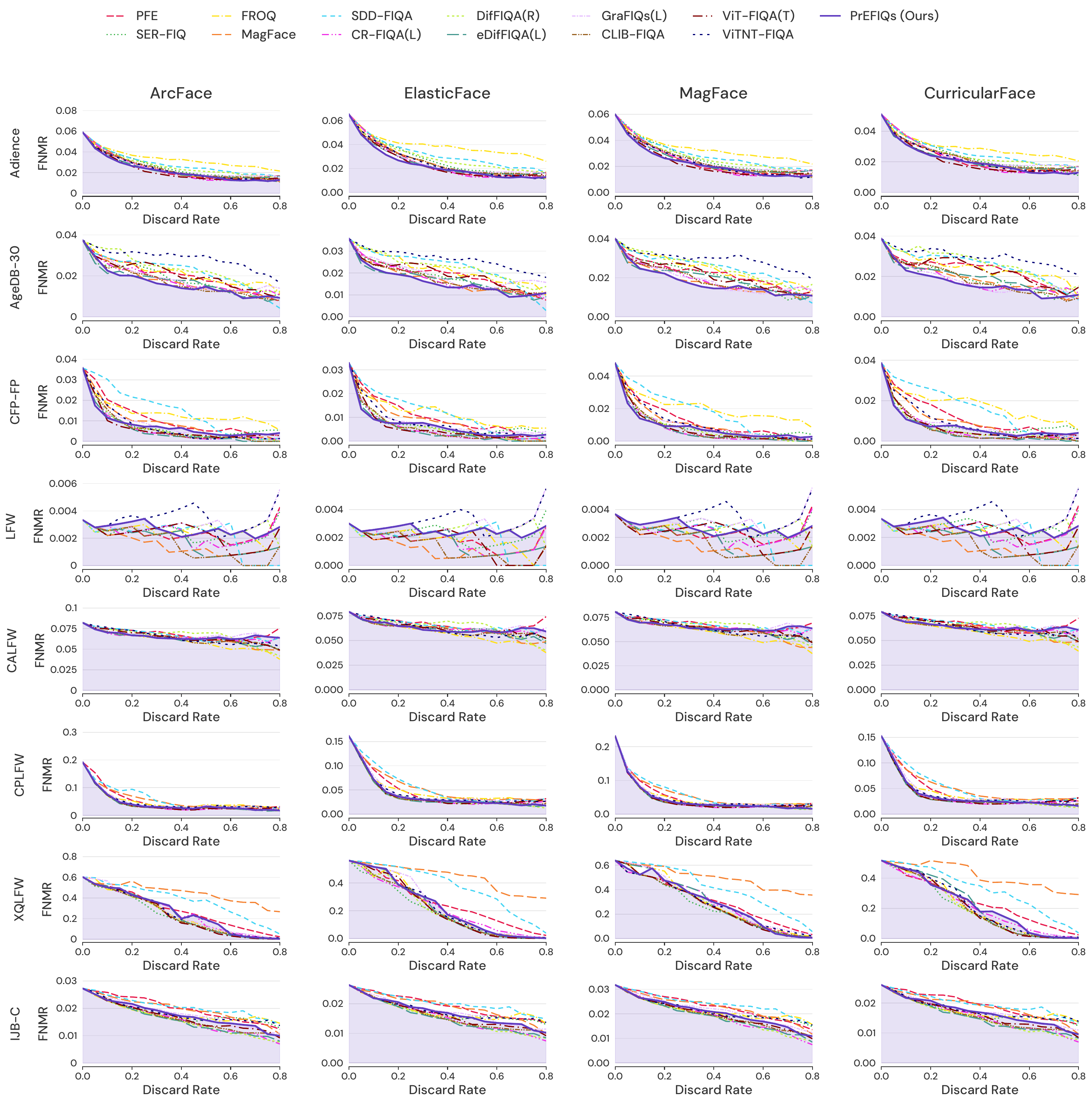}\vspace{-2mm}
\end{figure*}

\begin{figure*}
\caption{Comparison of EDC curves (FNMR at FMR=$1e^{-4}$) of PreFIQs against recent FIQA approaches.
    The results are shown for four FR models on eight benchmarks.
    Unsupervised approaches are visualized using dotted lines. Supervised methods are visualized with dashed lines.
    PreFIQs is visualized using a continous line with shaded AUC.
    For PreFIQs, unstructured $L_1$ magnitude pruning with $\rho=0.4$ is used.
    }
    \label{fig:edc_curves_supp_fnmr4}\vspace{-2mm}
    \centering
    \includegraphics[width=1.0\linewidth]{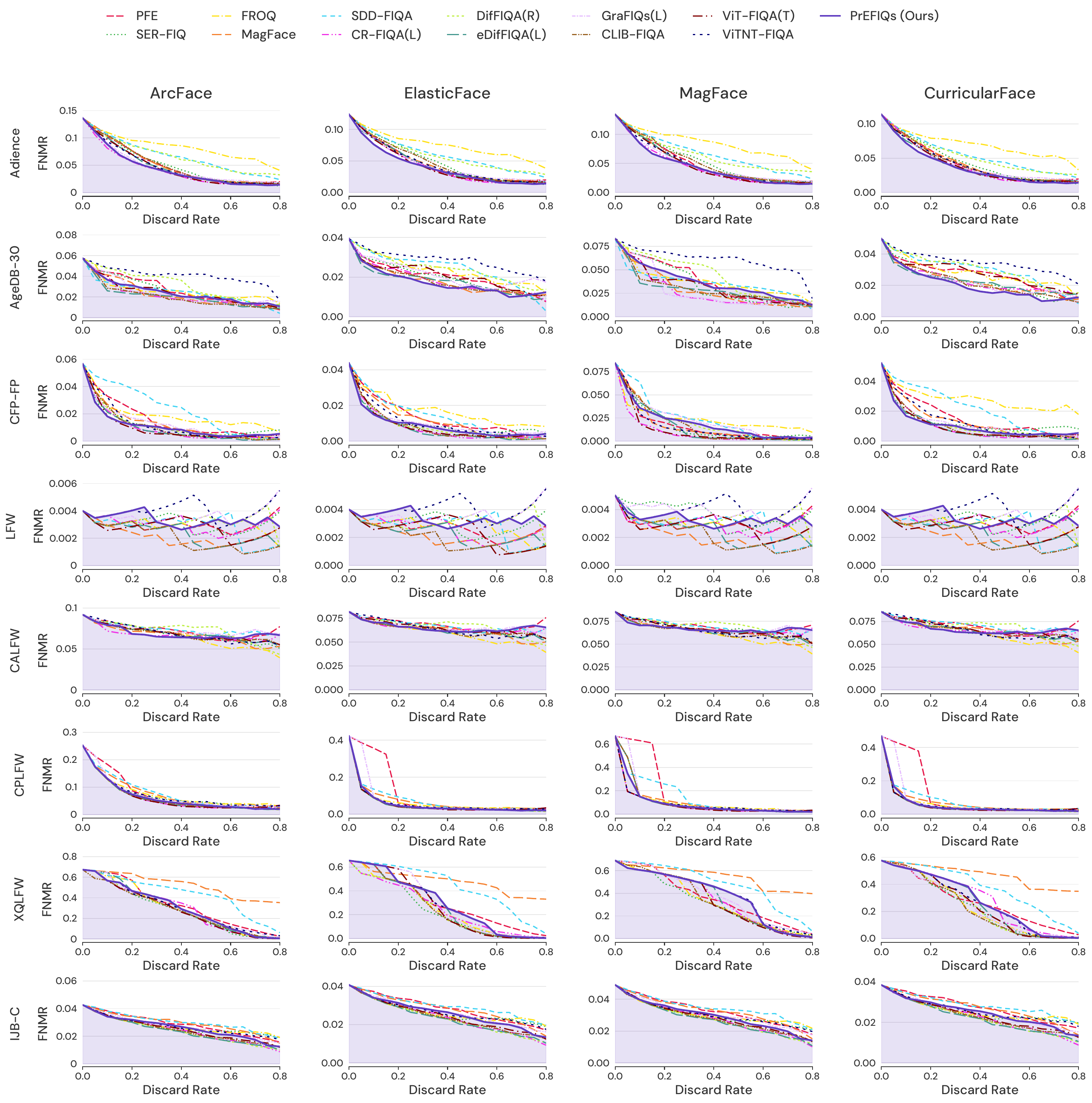}\vspace{-2mm}
\end{figure*}

This supplementary material sections contains the following supporting content:
\begin{itemize}
    \item Detailed pAUC results across all four evaluated FR models and pruning ratios $\rho$. These are provided for unstructured $L_1$ magnitude pruning (Table~\ref{tab:fiqa_resnet100_unstructured}), unstructured random pruning (Table~\ref{tab:fiqa_resnet100_random}), and structured pruning (Table~\ref{tab:fiqa_resnet100_structured}).
    \item A comprehensive comparison of FR verification accuracy across different pruning granularities (unstructured vs. structured) and parameter selection criteria (unstructured $L_1$ magnitude vs. unstructured random pruning), detailed in Table~\ref{tab:fr_pruning_comparison_full}.
    \item An extended comparison of PreFIQs (using unstructured $L_1$ magnitude pruning at $\rho=0.4$) against recent state-of-the-art FIQA approaches. Table~\ref{tab:sota_pauc_1e4} provides the pAUC results evaluated at an FMR of $10^{-4}$ to complement the results provided in the main paper.
    \item Error-Versus-Discard Characteristic (EDC) curves comparing PreFIQs ($\rho=0.4$) against recent FIQA methods. These curves are plotted for an FMR of $10^{-3}$ in Figure~\ref{fig:edc_curves_supp_fnmr3} and an FMR of $10^{-4}$ in Figure~\ref{fig:edc_curves_supp_fnmr4}.
    \item Additional evaluations utilizing a ResNet50 backbone. Table~\ref{tab:resnet50_fiqa_pauc_fnmr1e-3} presents the pAUC results at an FMR of $10^{-3}$ across all four FR models using unstructured $L_1$ magnitude pruning.
\end{itemize}

\begin{table}[!ht]
    \centering
\caption{Performance of \textbf{unstructured $L_1$ magnitude pruning} on four FR models using pAUC scores (discard rate = 0.3, FMR = $10^{-3}$). We exclude XQLFW from this average, as its quality labels were derived using SER-FIQ. The best pAUC value is shaded.}
    \resizebox{\columnwidth}{!}{%
\begin{NiceTabular}{c r |  r r r r r r r | r}
\Block{2-10}{\textbf{ArcFace~\cite{deng2019arcface}} - $pAUC * 10^{3} \, ($FMR$=10^{-3}) \, [\downarrow]$} \\
 \\
 {} & \textbf{{Methods}} & \textbf{Adience} & \textbf{AgeDB-30} & \textbf{CFP-FP} & \textbf{LFW} & \textbf{CALFW} & \textbf{CPLFW} & \textbf{XQLFW} & $\overline{pAUC}$ \\
\midrule
\Block[tikz={pattern = {Lines[angle=-45, distance=1.5mm, line width=0.5mm]}, pattern color=cyan!50}]{9-1} {} & \textbf{Unstructured $L_1$ Magnitude Pruning $\rho=$0.1} & $9.933$ & $6.866$ & $3.779$ & $0.849$ & $21.913$ & $\mathit{20.809}$ & $\mathit{139.497}$ & $10.691$ \\
{} & \textbf{Unstructured $L_1$ Magnitude Pruning $\rho=$0.2} & $10.088$ & $7.129$ & $\mathbf{3.479}$ & $0.912$ & $21.963$ & $\mathbf{20.761}$ & $\mathbf{137.626}$ & $10.722$ \\
{} & \textbf{Unstructured $L_1$ Magnitude Pruning $\rho=$0.3} & $9.798$ & $6.991$ & $3.972$ & $\mathbf{0.779}$ & $22.219$ & $21.045$ & $143.120$ & $10.801$ \\
{} & \textbf{Unstructured $L_1$ Magnitude Pruning $\rho=$0.4} & $10.009$ & $6.876$ & $3.755$ & $0.921$ & $\mathbf{20.979}$ & $21.180$ & $141.716$ & $\cellcolor{cyan!10}10.620$ \\
{} & \textbf{Unstructured $L_1$ Magnitude Pruning $\rho=$0.5} & $\mathbf{9.734}$ & $6.910$ & $\mathit{3.631}$ & $0.880$ & $21.502$ & $21.284$ & $143.139$ & $10.657$ \\
{} & \textbf{Unstructured $L_1$ Magnitude Pruning $\rho=$0.6} & $\mathit{9.765}$ & $\mathbf{6.699}$ & $3.832$ & $\mathit{0.788}$ & $\mathit{21.190}$ & $23.145$ & $146.722$ & $10.903$ \\
{} & \textbf{Unstructured $L_1$ Magnitude Pruning $\rho=$0.7} & $9.979$ & $\mathit{6.747}$ & $4.354$ & $0.873$ & $21.791$ & $26.264$ & $148.950$ & $11.668$ \\
{} & \textbf{Unstructured $L_1$ Magnitude Pruning $\rho=$0.8} & $11.466$ & $8.280$ & $8.139$ & $0.832$ & $22.271$ & $41.568$ & $160.413$ & $15.426$ \\
{} & \textbf{Unstructured $L_1$ Magnitude Pruning $\rho=$0.9} & $14.394$ & $8.932$ & $11.292$ & $0.856$ & $23.614$ & $56.227$ & $183.969$ & $19.219$ \\
\bottomrule

\Block{2-10}{\textbf{CurricularFace~\cite{curricularFace}} - $pAUC * 10^{3} \, ($FMR$=10^{-3}) \, [\downarrow]$} \\
 \\
 {} & \textbf{{Methods}} & \textbf{Adience} & \textbf{AgeDB-30} & \textbf{CFP-FP} & \textbf{LFW} & \textbf{CALFW} & \textbf{CPLFW} & \textbf{XQLFW} & $\overline{pAUC}$ \\
\midrule
\Block[tikz={pattern = {Lines[angle=-45, distance=1.5mm, line width=0.5mm]}, pattern color=cyan!50}]{9-1} {} & \textbf{Unstructured $L_1$ Magnitude Pruning $\rho=$0.1} & $8.986$ & $\mathit{7.018}$ & $3.732$ & $0.921$ & $21.154$ & $\mathit{17.853}$ & $124.686$ & $9.944$ \\
{} & \textbf{Unstructured $L_1$ Magnitude Pruning $\rho=$0.2} & $9.153$ & $7.593$ & $\mathbf{3.484}$ & $0.952$ & $21.252$ & $\mathbf{17.598}$ & $\mathbf{121.306}$ & $10.005$ \\
{} & \textbf{Unstructured $L_1$ Magnitude Pruning $\rho=$0.3} & $\mathit{8.842}$ & $7.032$ & $3.868$ & $\mathbf{0.779}$ & $21.441$ & $18.011$ & $124.430$ & $9.995$ \\
{} & \textbf{Unstructured $L_1$ Magnitude Pruning $\rho=$0.4} & $8.968$ & $7.020$ & $3.752$ & $0.921$ & $\mathit{20.577}$ & $18.239$ & $123.709$ & $9.913$ \\
{} & \textbf{Unstructured $L_1$ Magnitude Pruning $\rho=$0.5} & $\mathbf{8.786}$ & $7.345$ & $\mathit{3.554}$ & $0.892$ & $20.875$ & $18.171$ & $124.025$ & $9.937$ \\
{} & \textbf{Unstructured $L_1$ Magnitude Pruning $\rho=$0.6} & $8.882$ & $\mathbf{6.927}$ & $3.864$ & $\mathit{0.808}$ & $\mathbf{20.554}$ & $18.380$ & $\mathit{123.356}$ & $\cellcolor{cyan!10}9.903$ \\
{} & \textbf{Unstructured $L_1$ Magnitude Pruning $\rho=$0.7} & $9.058$ & $7.333$ & $4.493$ & $0.873$ & $21.031$ & $22.024$ & $133.094$ & $10.802$ \\
{} & \textbf{Unstructured $L_1$ Magnitude Pruning $\rho=$0.8} & $10.412$ & $8.578$ & $8.283$ & $0.832$ & $21.529$ & $33.501$ & $145.322$ & $13.856$ \\
{} & \textbf{Unstructured $L_1$ Magnitude Pruning $\rho=$0.9} & $13.144$ & $9.495$ & $11.581$ & $0.856$ & $22.612$ & $45.848$ & $162.629$ & $17.256$ \\
\bottomrule

\Block{2-10}{\textbf{ElasticFace~\cite{elasticface}} - $pAUC * 10^{3} \, ($FMR$=10^{-3}) \, [\downarrow]$} \\
 \\
 {} & \textbf{{Methods}} & \textbf{Adience} & \textbf{AgeDB-30} & \textbf{CFP-FP} & \textbf{LFW} & \textbf{CALFW} & \textbf{CPLFW} & \textbf{XQLFW} & $\overline{pAUC}$ \\
\midrule
\Block[tikz={pattern = {Lines[angle=-45, distance=1.5mm, line width=0.5mm]}, pattern color=cyan!50}]{9-1} {} & \textbf{Unstructured $L_1$ Magnitude Pruning $\rho=$0.1} & $10.651$ & $\mathbf{6.314}$ & $\mathit{3.201}$ & $0.785$ & $21.125$ & $\mathit{19.664}$ & $131.822$ & $10.290$ \\
{} & \textbf{Unstructured $L_1$ Magnitude Pruning $\rho=$0.2} & $10.954$ & $6.958$ & $3.229$ & $0.796$ & $21.144$ & $\mathbf{19.391}$ & $\mathit{129.852}$ & $10.412$ \\
{} & \textbf{Unstructured $L_1$ Magnitude Pruning $\rho=$0.3} & $\mathit{10.592}$ & $\mathit{6.586}$ & $3.396$ & $\mathbf{0.664}$ & $21.356$ & $19.740$ & $\mathbf{129.723}$ & $10.389$ \\
{} & \textbf{Unstructured $L_1$ Magnitude Pruning $\rho=$0.4} & $10.664$ & $6.600$ & $3.287$ & $0.805$ & $\mathit{20.324}$ & $19.926$ & $135.726$ & $\cellcolor{cyan!10}10.268$ \\
{} & \textbf{Unstructured $L_1$ Magnitude Pruning $\rho=$0.5} & $\mathbf{10.584}$ & $6.657$ & $\mathbf{3.102}$ & $0.776$ & $20.690$ & $19.889$ & $139.316$ & $10.283$ \\
{} & \textbf{Unstructured $L_1$ Magnitude Pruning $\rho=$0.6} & $10.662$ & $6.667$ & $3.358$ & $0.725$ & $\mathbf{20.318}$ & $20.120$ & $148.875$ & $10.308$ \\
{} & \textbf{Unstructured $L_1$ Magnitude Pruning $\rho=$0.7} & $11.088$ & $6.686$ & $3.879$ & $0.757$ & $20.884$ & $23.476$ & $146.510$ & $11.128$ \\
{} & \textbf{Unstructured $L_1$ Magnitude Pruning $\rho=$0.8} & $13.191$ & $7.920$ & $7.247$ & $\mathit{0.715}$ & $21.358$ & $35.078$ & $159.555$ & $14.252$ \\
{} & \textbf{Unstructured $L_1$ Magnitude Pruning $\rho=$0.9} & $16.575$ & $9.039$ & $10.293$ & $0.739$ & $22.483$ & $48.956$ & $172.481$ & $18.014$ \\
\bottomrule

\Block{2-10}{\textbf{MagFace~\cite{MagFace}} - $pAUC * 10^{3} \, ($FMR$=10^{-3}) \, [\downarrow]$} \\
 \\
 {} & \textbf{{Methods}} & \textbf{Adience} & \textbf{AgeDB-30} & \textbf{CFP-FP} & \textbf{LFW} & \textbf{CALFW} & \textbf{CPLFW} & \textbf{XQLFW} & $\overline{pAUC}$ \\
\midrule
\Block[tikz={pattern = {Lines[angle=-45, distance=1.5mm, line width=0.5mm]}, pattern color=cyan!50}]{9-1} {} & \textbf{Unstructured $L_1$ Magnitude Pruning $\rho=$0.1} & $10.128$ & $\mathbf{7.040}$ & $\mathbf{4.672}$ & $0.910$ & $21.547$ & $23.441$ & $152.317$ & $11.290$ \\
{} & \textbf{Unstructured $L_1$ Magnitude Pruning $\rho=$0.2} & $10.299$ & $7.999$ & $\mathit{4.683}$ & $0.921$ & $21.637$ & $\mathbf{22.645}$ & $\mathbf{148.797}$ & $11.364$ \\
{} & \textbf{Unstructured $L_1$ Magnitude Pruning $\rho=$0.3} & $9.979$ & $7.236$ & $4.924$ & $\mathbf{0.788}$ & $21.880$ & $\mathit{23.096}$ & $\mathit{150.416}$ & $11.317$ \\
{} & \textbf{Unstructured $L_1$ Magnitude Pruning $\rho=$0.4} & $10.146$ & $7.432$ & $4.805$ & $0.947$ & $\mathit{20.981}$ & $23.261$ & $154.940$ & $11.262$ \\
{} & \textbf{Unstructured $L_1$ Magnitude Pruning $\rho=$0.5} & $\mathit{9.894}$ & $7.285$ & $4.749$ & $0.936$ & $21.258$ & $23.174$ & $153.128$ & $\cellcolor{cyan!10}11.216$ \\
{} & \textbf{Unstructured $L_1$ Magnitude Pruning $\rho=$0.6} & $\mathbf{9.881}$ & $\mathit{7.177}$ & $5.053$ & $\mathit{0.834}$ & $\mathbf{20.826}$ & $23.531$ & $157.902$ & $11.217$ \\
{} & \textbf{Unstructured $L_1$ Magnitude Pruning $\rho=$0.7} & $10.115$ & $7.499$ & $5.591$ & $0.917$ & $21.640$ & $29.432$ & $171.570$ & $12.532$ \\
{} & \textbf{Unstructured $L_1$ Magnitude Pruning $\rho=$0.8} & $11.667$ & $9.018$ & $9.489$ & $0.841$ & $22.207$ & $54.756$ & $179.703$ & $17.996$ \\
{} & \textbf{Unstructured $L_1$ Magnitude Pruning $\rho=$0.9} & $14.919$ & $10.391$ & $13.723$ & $1.016$ & $23.270$ & $81.149$ & $192.868$ & $24.078$ \\
\bottomrule
\end{NiceTabular}
}
\label{tab:fiqa_resnet100_unstructured}
\end{table}

\begin{table}[!ht]
    \centering
\caption{Performance of \textbf{unstructured random pruning} on four FR models using pAUC scores (discard rate = 0.3, FMR = $10^{-3}$). We exclude XQLFW from this average, as its quality labels were derived using SER-FIQ. The best pAUC value is shaded.}
    \resizebox{\columnwidth}{!}{%
\begin{NiceTabular}{c r |  r r r r r r r | r}
\Block{2-10}{\textbf{ArcFace~\cite{deng2019arcface}} - $pAUC * 10^{3} \, ($FMR$=10^{-3}) \, [\downarrow]$} \\
 \\
 {} & \textbf{{Methods}} & \textbf{Adience} & \textbf{AgeDB-30} & \textbf{CFP-FP} & \textbf{LFW} & \textbf{CALFW} & \textbf{CPLFW} & \textbf{XQLFW} & $\overline{pAUC}$ \\
\midrule
\Block[tikz={pattern = {Hatch[angle=45, distance=1.5mm, line width=0.5mm]}, pattern color=violet!40}]{9-1} {} & \textbf{Unstructured Random Pruning $\rho=$0.1} & $\mathbf{15.057}$ & $\mathbf{8.802}$ & $\mathbf{10.271}$ & $\mathbf{0.738}$ & $\mathbf{22.996}$ & $\mathbf{50.725}$ & $185.933$ & $\cellcolor{violet!10}18.098$ \\
{} & \textbf{Unstructured Random Pruning $\rho=$0.2} & $17.838$ & $\mathit{9.569}$ & $12.682$ & $1.122$ & $24.483$ & $60.280$ & $187.807$ & $20.996$ \\
{} & \textbf{Unstructured Random Pruning $\rho=$0.3} & $16.583$ & $10.460$ & $12.331$ & $0.912$ & $24.369$ & $60.005$ & $\mathbf{180.625}$ & $20.777$ \\
{} & \textbf{Unstructured Random Pruning $\rho=$0.4} & $16.307$ & $10.440$ & $12.339$ & $0.892$ & $\mathit{24.160}$ & $60.143$ & $\mathit{183.353}$ & $20.713$ \\
{} & \textbf{Unstructured Random Pruning $\rho=$0.5} & $\mathit{15.748}$ & $10.466$ & $12.097$ & $0.925$ & $24.225$ & $60.499$ & $185.443$ & $20.660$ \\
{} & \textbf{Unstructured Random Pruning $\rho=$0.6} & $15.963$ & $10.769$ & $\mathit{12.005}$ & $0.917$ & $24.687$ & $\mathit{58.891}$ & $184.489$ & $20.539$ \\
{} & \textbf{Unstructured Random Pruning $\rho=$0.7} & $16.281$ & $10.393$ & $12.139$ & $0.991$ & $24.426$ & $59.896$ & $183.915$ & $20.688$ \\
{} & \textbf{Unstructured Random Pruning $\rho=$0.8} & $17.055$ & $10.729$ & $12.192$ & $0.908$ & $24.665$ & $59.558$ & $185.973$ & $20.851$ \\
{} & \textbf{Unstructured Random Pruning $\rho=$0.9} & $16.982$ & $10.296$ & $12.013$ & $\mathit{0.891}$ & $24.739$ & $59.662$ & $186.222$ & $20.764$ \\
\bottomrule

\Block{2-10}{\textbf{CurricularFace~\cite{curricularFace}} - $pAUC * 10^{3} \, ($FMR$=10^{-3}) \, [\downarrow]$} \\
 \\
 {} & \textbf{{Methods}} & \textbf{Adience} & \textbf{AgeDB-30} & \textbf{CFP-FP} & \textbf{LFW} & \textbf{CALFW} & \textbf{CPLFW} & \textbf{XQLFW} & $\overline{pAUC}$ \\
\midrule
\Block[tikz={pattern = {Hatch[angle=45, distance=1.5mm, line width=0.5mm]}, pattern color=violet!40}]{9-1} {} & \textbf{Unstructured Random Pruning $\rho=$0.1} & $\mathbf{13.073}$ & $\mathbf{10.041}$ & $\mathbf{10.247}$ & $\mathbf{0.738}$ & $\mathbf{22.443}$ & $\mathbf{42.386}$ & $166.993$ & $\cellcolor{violet!10}16.488$ \\
{} & \textbf{Unstructured Random Pruning $\rho=$0.2} & $15.459$ & $\mathit{10.792}$ & $12.186$ & $1.122$ & $23.692$ & $47.304$ & $166.937$ & $18.426$ \\
{} & \textbf{Unstructured Random Pruning $\rho=$0.3} & $14.480$ & $11.283$ & $11.765$ & $0.912$ & $23.264$ & $49.333$ & $162.053$ & $18.506$ \\
{} & \textbf{Unstructured Random Pruning $\rho=$0.4} & $14.282$ & $11.352$ & $11.831$ & $0.892$ & $\mathit{23.069}$ & $46.203$ & $\mathit{161.380}$ & $17.938$ \\
{} & \textbf{Unstructured Random Pruning $\rho=$0.5} & $\mathit{14.103}$ & $11.350$ & $11.629$ & $0.925$ & $23.155$ & $47.227$ & $163.548$ & $18.065$ \\
{} & \textbf{Unstructured Random Pruning $\rho=$0.6} & $14.334$ & $11.735$ & $11.667$ & $0.917$ & $23.542$ & $48.328$ & $\mathbf{158.221}$ & $18.420$ \\
{} & \textbf{Unstructured Random Pruning $\rho=$0.7} & $14.583$ & $11.140$ & $11.578$ & $0.991$ & $23.213$ & $46.416$ & $162.351$ & $17.987$ \\
{} & \textbf{Unstructured Random Pruning $\rho=$0.8} & $15.101$ & $11.521$ & $11.629$ & $0.908$ & $23.502$ & $46.217$ & $164.159$ & $18.147$ \\
{} & \textbf{Unstructured Random Pruning $\rho=$0.9} & $15.174$ & $11.329$ & $\mathit{11.404}$ & $\mathit{0.891}$ & $23.603$ & $\mathit{46.027}$ & $164.509$ & $18.071$ \\
\bottomrule

\Block{2-10}{\textbf{ElasticFace~\cite{elasticface}} - $pAUC * 10^{3} \, ($FMR$=10^{-3}) \, [\downarrow]$} \\
 \\
 {} & \textbf{{Methods}} & \textbf{Adience} & \textbf{AgeDB-30} & \textbf{CFP-FP} & \textbf{LFW} & \textbf{CALFW} & \textbf{CPLFW} & \textbf{XQLFW} & $\overline{pAUC}$ \\
\midrule
\Block[tikz={pattern = {Hatch[angle=45, distance=1.5mm, line width=0.5mm]}, pattern color=violet!40}]{9-1} {} & \textbf{Unstructured Random Pruning $\rho=$0.1} & $\mathbf{17.014}$ & $\mathbf{8.479}$ & $\mathbf{8.921}$ & $\mathbf{0.621}$ & $\mathbf{22.330}$ & $\mathbf{44.025}$ & $175.487$ & $\cellcolor{violet!10}16.898$ \\
{} & \textbf{Unstructured Random Pruning $\rho=$0.2} & $19.909$ & $\mathit{9.445}$ & $11.197$ & $1.005$ & $23.520$ & $51.749$ & $176.004$ & $19.471$ \\
{} & \textbf{Unstructured Random Pruning $\rho=$0.3} & $18.670$ & $9.683$ & $10.557$ & $0.796$ & $23.383$ & $51.002$ & $\mathbf{170.618}$ & $19.015$ \\
{} & \textbf{Unstructured Random Pruning $\rho=$0.4} & $18.352$ & $9.976$ & $10.935$ & $0.776$ & $\mathit{23.089}$ & $51.820$ & $\mathit{170.678}$ & $19.158$ \\
{} & \textbf{Unstructured Random Pruning $\rho=$0.5} & $\mathit{17.806}$ & $9.854$ & $10.513$ & $0.809$ & $23.295$ & $51.880$ & $173.129$ & $19.026$ \\
{} & \textbf{Unstructured Random Pruning $\rho=$0.6} & $18.054$ & $10.289$ & $10.548$ & $0.854$ & $23.671$ & $\mathit{50.339}$ & $171.993$ & $18.959$ \\
{} & \textbf{Unstructured Random Pruning $\rho=$0.7} & $18.405$ & $9.777$ & $10.423$ & $0.874$ & $23.376$ & $51.147$ & $171.672$ & $19.000$ \\
{} & \textbf{Unstructured Random Pruning $\rho=$0.8} & $19.179$ & $10.138$ & $10.359$ & $0.792$ & $23.619$ & $50.854$ & $173.737$ & $19.157$ \\
{} & \textbf{Unstructured Random Pruning $\rho=$0.9} & $19.155$ & $9.768$ & $\mathit{10.309}$ & $\mathit{0.775}$ & $23.754$ & $51.312$ & $173.965$ & $19.179$ \\
\bottomrule

\Block{2-10}{\textbf{MagFace~\cite{MagFace}} - $pAUC * 10^{3} \, ($FMR$=10^{-3}) \, [\downarrow]$} \\
 \\
 {} & \textbf{{Methods}} & \textbf{Adience} & \textbf{AgeDB-30} & \textbf{CFP-FP} & \textbf{LFW} & \textbf{CALFW} & \textbf{CPLFW} & \textbf{XQLFW} & $\overline{pAUC}$ \\
\midrule
\Block[tikz={pattern = {Hatch[angle=45, distance=1.5mm, line width=0.5mm]}, pattern color=violet!40}]{9-1} {} & \textbf{Unstructured Random Pruning $\rho=$0.1} & $\mathbf{15.261}$ & $\mathbf{9.712}$ & $\mathbf{12.488}$ & $\mathbf{0.822}$ & $\mathbf{23.015}$ & $\mathbf{68.953}$ & $196.552$ & $\cellcolor{violet!10}21.708$ \\
{} & \textbf{Unstructured Random Pruning $\rho=$0.2} & $18.076$ & $\mathit{10.819}$ & $15.366$ & $1.292$ & $24.345$ & $\mathit{87.547}$ & $197.676$ & $26.241$ \\
{} & \textbf{Unstructured Random Pruning $\rho=$0.3} & $17.095$ & $11.352$ & $14.751$ & $0.992$ & $24.040$ & $87.775$ & $\mathit{193.409}$ & $26.001$ \\
{} & \textbf{Unstructured Random Pruning $\rho=$0.4} & $16.734$ & $11.402$ & $15.136$ & $0.972$ & $\mathit{23.896}$ & $88.859$ & $193.840$ & $26.167$ \\
{} & \textbf{Unstructured Random Pruning $\rho=$0.5} & $\mathit{16.289}$ & $11.336$ & $14.845$ & $1.005$ & $23.924$ & $89.931$ & $195.817$ & $26.222$ \\
{} & \textbf{Unstructured Random Pruning $\rho=$0.6} & $16.681$ & $11.614$ & $14.803$ & $0.997$ & $24.359$ & $87.955$ & $\mathbf{193.222}$ & $26.068$ \\
{} & \textbf{Unstructured Random Pruning $\rho=$0.7} & $17.001$ & $11.159$ & $14.704$ & $1.069$ & $24.148$ & $88.618$ & $194.451$ & $26.117$ \\
{} & \textbf{Unstructured Random Pruning $\rho=$0.8} & $17.618$ & $11.563$ & $14.634$ & $0.989$ & $24.376$ & $88.542$ & $196.422$ & $26.287$ \\
{} & \textbf{Unstructured Random Pruning $\rho=$0.9} & $17.572$ & $11.068$ & $\mathit{14.375}$ & $\mathit{0.970}$ & $24.555$ & $89.114$ & $196.850$ & $26.276$ \\
\bottomrule
\end{NiceTabular}
}
\label{tab:fiqa_resnet100_random}
\end{table}

\begin{table}[!ht]
    \centering
\caption{Performance of \textbf{structured pruning} on four FR models using pAUC scores (discard rate = 0.3, FMR = $10^{-3}$). We exclude XQLFW from this average, as its quality labels were derived using SER-FIQ. The best pAUC value is shaded.}
    \resizebox{\columnwidth}{!}{%
\begin{NiceTabular}{c r |  r r r r r r r | r}
\Block{2-10}{\textbf{ArcFace~\cite{deng2019arcface}} - $pAUC * 10^{3} \, ($FMR$=10^{-3}) \, [\downarrow]$} \\
 \\
 {} & \textbf{{Methods}} & \textbf{Adience} & \textbf{AgeDB-30} & \textbf{CFP-FP} & \textbf{LFW} & \textbf{CALFW} & \textbf{CPLFW} & \textbf{XQLFW} & $\overline{pAUC}$ \\
\midrule
\Block[tikz={pattern = {Dots[angle=45, distance=1.5mm, radius=0.3mm]}, pattern color=orange!50}]{9-1} {} & \textbf{Structured Pruning $\rho=$0.1} & $\mathbf{10.682}$ & $\mathbf{8.092}$ & $\mathbf{4.214}$ & $\mathbf{0.798}$ & $\mathbf{21.534}$ & $\mathbf{22.337}$ & $\mathbf{146.636}$ & $\cellcolor{orange!10}11.276$ \\
{} & \textbf{Structured Pruning $\rho=$0.2} & $\mathit{11.866}$ & $\mathit{9.480}$ & $\mathit{8.575}$ & $0.964$ & $\mathit{22.461}$ & $\mathit{48.433}$ & $\mathit{160.614}$ & $16.963$ \\
{} & \textbf{Structured Pruning $\rho=$0.3} & $16.136$ & $10.348$ & $11.902$ & $1.170$ & $23.726$ & $58.509$ & $174.307$ & $20.299$ \\
{} & \textbf{Structured Pruning $\rho=$0.4} & $16.447$ & $11.296$ & $12.291$ & $0.932$ & $25.109$ & $55.601$ & $174.252$ & $20.279$ \\
{} & \textbf{Structured Pruning $\rho=$0.5} & $16.254$ & $11.354$ & $11.728$ & $\mathit{0.909}$ & $24.631$ & $57.294$ & $176.949$ & $20.362$ \\
{} & \textbf{Structured Pruning $\rho=$0.6} & $15.875$ & $11.021$ & $11.655$ & $0.929$ & $24.645$ & $57.736$ & $177.852$ & $20.310$ \\
{} & \textbf{Structured Pruning $\rho=$0.7} & $17.205$ & $10.568$ & $11.681$ & $1.005$ & $24.748$ & $57.508$ & $173.673$ & $20.452$ \\
{} & \textbf{Structured Pruning $\rho=$0.8} & $16.181$ & $10.400$ & $11.986$ & $1.011$ & $24.774$ & $58.010$ & $175.532$ & $20.394$ \\
{} & \textbf{Structured Pruning $\rho=$0.9} & $16.529$ & $9.951$ & $12.482$ & $0.967$ & $24.638$ & $59.317$ & $178.910$ & $20.647$ \\
\bottomrule

\Block{2-10}{\textbf{CurricularFace~\cite{curricularFace}} - $pAUC * 10^{3} \, ($FMR$=10^{-3}) \, [\downarrow]$} \\
 \\
 {} & \textbf{{Methods}} & \textbf{Adience} & \textbf{AgeDB-30} & \textbf{CFP-FP} & \textbf{LFW} & \textbf{CALFW} & \textbf{CPLFW} & \textbf{XQLFW} & $\overline{pAUC}$ \\
\midrule
\Block[tikz={pattern = {Dots[angle=45, distance=1.5mm, radius=0.3mm]}, pattern color=orange!50}]{9-1} {} & \textbf{Structured Pruning $\rho=$0.1} & $\mathbf{9.438}$ & $\mathbf{8.264}$ & $\mathbf{4.576}$ & $\mathbf{0.851}$ & $\mathbf{20.940}$ & $\mathbf{19.171}$ & $\mathbf{133.453}$ & $\cellcolor{orange!10}10.540$ \\
{} & \textbf{Structured Pruning $\rho=$0.2} & $\mathit{10.507}$ & $\mathit{10.131}$ & $\mathit{9.091}$ & $0.964$ & $\mathit{21.591}$ & $\mathit{39.515}$ & $\mathit{139.968}$ & $15.300$ \\
{} & \textbf{Structured Pruning $\rho=$0.3} & $14.202$ & $10.617$ & $12.293$ & $1.170$ & $22.872$ & $48.333$ & $158.205$ & $18.248$ \\
{} & \textbf{Structured Pruning $\rho=$0.4} & $14.424$ & $12.192$ & $12.186$ & $0.932$ & $23.906$ & $47.932$ & $156.167$ & $18.595$ \\
{} & \textbf{Structured Pruning $\rho=$0.5} & $14.431$ & $11.997$ & $11.473$ & $\mathit{0.909}$ & $23.468$ & $47.403$ & $155.834$ & $18.280$ \\
{} & \textbf{Structured Pruning $\rho=$0.6} & $14.240$ & $11.372$ & $11.365$ & $0.929$ & $23.537$ & $46.968$ & $156.709$ & $18.069$ \\
{} & \textbf{Structured Pruning $\rho=$0.7} & $15.091$ & $10.823$ & $11.632$ & $1.005$ & $23.723$ & $47.080$ & $157.546$ & $18.226$ \\
{} & \textbf{Structured Pruning $\rho=$0.8} & $14.372$ & $11.092$ & $11.446$ & $1.011$ & $23.659$ & $47.229$ & $155.716$ & $18.135$ \\
{} & \textbf{Structured Pruning $\rho=$0.9} & $14.664$ & $10.698$ & $11.942$ & $0.967$ & $23.536$ & $48.810$ & $162.417$ & $18.436$ \\
\bottomrule

\Block{2-10}{\textbf{ElasticFace~\cite{elasticface}} - $pAUC * 10^{3} \, ($FMR$=10^{-3}) \, [\downarrow]$} \\
 \\
 {} & \textbf{{Methods}} & \textbf{Adience} & \textbf{AgeDB-30} & \textbf{CFP-FP} & \textbf{LFW} & \textbf{CALFW} & \textbf{CPLFW} & \textbf{XQLFW} & $\overline{pAUC}$ \\
\midrule
\Block[tikz={pattern = {Dots[angle=45, distance=1.5mm, radius=0.3mm]}, pattern color=orange!50}]{9-1} {} & \textbf{Structured Pruning $\rho=$0.1} & $\mathbf{11.861}$ & $\mathbf{7.862}$ & $\mathbf{3.872}$ & $\mathbf{0.692}$ & $\mathbf{20.568}$ & $\mathbf{20.907}$ & $\mathbf{138.529}$ & $\cellcolor{orange!10}10.960$ \\
{} & \textbf{Structured Pruning $\rho=$0.2} & $\mathit{13.507}$ & $\mathit{9.249}$ & $\mathit{8.052}$ & $0.846$ & $\mathit{21.444}$ & $\mathit{41.140}$ & $\mathit{156.850}$ & $15.706$ \\
{} & \textbf{Structured Pruning $\rho=$0.3} & $18.070$ & $9.691$ & $10.516$ & $1.053$ & $22.736$ & $50.392$ & $167.648$ & $18.743$ \\
{} & \textbf{Structured Pruning $\rho=$0.4} & $18.709$ & $10.760$ & $10.807$ & $0.815$ & $24.068$ & $49.996$ & $165.012$ & $19.192$ \\
{} & \textbf{Structured Pruning $\rho=$0.5} & $18.484$ & $10.698$ & $10.304$ & $\mathit{0.793}$ & $23.627$ & $49.438$ & $164.708$ & $18.891$ \\
{} & \textbf{Structured Pruning $\rho=$0.6} & $18.115$ & $10.198$ & $10.425$ & $0.813$ & $23.675$ & $49.046$ & $165.614$ & $18.712$ \\
{} & \textbf{Structured Pruning $\rho=$0.7} & $19.485$ & $9.707$ & $10.470$ & $0.890$ & $23.947$ & $49.210$ & $165.305$ & $18.952$ \\
{} & \textbf{Structured Pruning $\rho=$0.8} & $18.584$ & $9.732$ & $10.521$ & $0.906$ & $23.850$ & $49.320$ & $168.750$ & $18.819$ \\
{} & \textbf{Structured Pruning $\rho=$0.9} & $18.753$ & $9.396$ & $10.646$ & $0.851$ & $23.601$ & $50.980$ & $171.624$ & $19.038$ \\
\bottomrule

\Block{2-10}{\textbf{MagFace~\cite{MagFace}} - $pAUC * 10^{3} \, ($FMR$=10^{-3}) \, [\downarrow]$} \\
 \\
 {} & \textbf{{Methods}} & \textbf{Adience} & \textbf{AgeDB-30} & \textbf{CFP-FP} & \textbf{LFW} & \textbf{CALFW} & \textbf{CPLFW} & \textbf{XQLFW} & $\overline{pAUC}$ \\
\midrule
\Block[tikz={pattern = {Dots[angle=45, distance=1.5mm, radius=0.3mm]}, pattern color=orange!50}]{9-1} {} & \textbf{Structured Pruning $\rho=$0.1} & $\mathbf{10.798}$ & $\mathbf{8.508}$ & $\mathbf{5.147}$ & $\mathbf{0.824}$ & $\mathbf{21.508}$ & $\mathbf{23.844}$ & $\mathbf{158.485}$ & $\cellcolor{orange!10}11.771$ \\
{} & \textbf{Structured Pruning $\rho=$0.2} & $\mathit{12.385}$ & $\mathit{10.379}$ & $\mathit{11.155}$ & $1.052$ & $\mathit{22.203}$ & $\mathit{68.994}$ & $\mathit{171.634}$ & $21.028$ \\
{} & \textbf{Structured Pruning $\rho=$0.3} & $16.519$ & $11.074$ & $15.199$ & $1.307$ & $23.296$ & $85.109$ & $179.578$ & $25.417$ \\
{} & \textbf{Structured Pruning $\rho=$0.4} & $17.022$ & $12.235$ & $15.101$ & $0.993$ & $24.704$ & $86.582$ & $186.718$ & $26.106$ \\
{} & \textbf{Structured Pruning $\rho=$0.5} & $16.890$ & $12.097$ & $14.587$ & $\mathit{0.988}$ & $24.281$ & $86.486$ & $187.859$ & $25.888$ \\
{} & \textbf{Structured Pruning $\rho=$0.6} & $16.528$ & $11.688$ & $14.496$ & $1.009$ & $24.339$ & $82.950$ & $187.531$ & $25.168$ \\
{} & \textbf{Structured Pruning $\rho=$0.7} & $17.828$ & $11.149$ & $14.474$ & $1.085$ & $24.584$ & $85.853$ & $187.248$ & $25.829$ \\
{} & \textbf{Structured Pruning $\rho=$0.8} & $16.881$ & $11.296$ & $14.667$ & $1.091$ & $24.497$ & $86.012$ & $191.370$ & $25.741$ \\
{} & \textbf{Structured Pruning $\rho=$0.9} & $17.257$ & $10.736$ & $15.155$ & $1.064$ & $24.268$ & $87.703$ & $194.474$ & $26.031$ \\
\bottomrule
\end{NiceTabular}
}
\label{tab:fiqa_resnet100_structured}
\end{table}

\begin{table}[!ht]
    \centering
    \caption{Verification accuracy (\%) of ResNet-100 under different pruning strategies at rates $\rho \in \{0.1, \ldots, 0.9\}$. The \textbf{best} and \textit{second-best} results per dataset are highlighted.}
    \resizebox{\columnwidth}{!}{%
    \begin{NiceTabular}{c r |  r r r r r r | r}
    \Block{2-9}{\textbf{Granularity of Pruning - Comparison between unstructured and structured model pruning} $[\uparrow]$} \\
    \\
    {} & \textbf{Methods} & \textbf{LFW} & \textbf{CFP-FP} & \textbf{CFP-FF} & \textbf{AgeDB-30} & \textbf{CALFW} & \textbf{CPLFW} & $\overline{\text{Acc}}$ [$\uparrow$] \\
    \midrule
    \Block[tikz={pattern = {Lines[angle=-45, distance=1.5mm, line width=0.5mm]}, pattern color=gray!50}]{1-1} {} & \textbf{ResNet-100 (unpruned)} & $\mathbf{99.80}$ & $\mathit{96.67}$ & $\mathit{99.89}$ & $98.35$ & $\mathit{96.15}$ & $\mathbf{93.32}$ & $\cellcolor{gray!30}97.36$ \\
    \dashmidrule
    \Block[tikz={pattern = {Lines[angle=-45, distance=1.5mm, line width=0.5mm]}, pattern color=cyan!50}]{9-1} {} & \textbf{Unstructured\ $\rho{=}0.1$} & $\mathbf{99.80}$ & $\mathit{96.67}$ & $\mathit{99.89}$ & $\mathbf{98.43}$ & $\mathbf{96.17}$ & $\mathit{93.23}$ & $97.37$ \\
    {} & \textbf{Unstructured\ $\rho{=}0.2$} & $\mathbf{99.80}$ & $\mathbf{96.70}$ & $\mathit{99.89}$ & $\mathbf{98.43}$ & $\mathbf{96.17}$ & $\mathit{93.23}$ & $\cellcolor{cyan!10}97.37$ \\
    {} & \textbf{Unstructured\ $\rho{=}0.3$} & $\mathbf{99.80}$ & $96.59$ & $\mathit{99.89}$ & $\mathit{98.42}$ & $96.08$ & $93.17$ & $97.32$ \\
    {} & \textbf{Unstructured\ $\rho{=}0.4$} & $\mathbf{99.80}$ & $96.44$ & $\mathit{99.89}$ & $98.32$ & $96.08$ & $93.18$ & $97.29$ \\
    {} & \textbf{Unstructured\ $\rho{=}0.5$} & $\mathbf{99.80}$ & $96.36$ & $\mathbf{99.90}$ & $98.20$ & $96.00$ & $92.70$ & $97.16$ \\
    {} & \textbf{Unstructured\ $\rho{=}0.6$} & $\mathbf{99.80}$ & $95.79$ & $\mathbf{99.90}$ & $98.05$ & $96.02$ & $92.22$ & $96.96$ \\
    {} & \textbf{Unstructured\ $\rho{=}0.7$} & $99.75$ & $94.87$ & $99.79$ & $97.50$ & $95.85$ & $90.38$ & $96.36$ \\
    {} & \textbf{Unstructured\ $\rho{=}0.8$} & $99.57$ & $87.50$ & $99.39$ & $94.70$ & $94.35$ & $82.92$ & $93.07$ \\
    {} & \textbf{Unstructured\ $\rho{=}0.9$} & $90.45$ & $65.76$ & $91.87$ & $75.27$ & $78.37$ & $58.87$ & $76.76$ \\
    \dashmidrule
    \Block[tikz={pattern = {Dots[angle=45, distance=1.5mm, radius=0.3mm]}, pattern color=orange!50}]{9-1} {} & \textbf{Structured\ $\rho{=}0.1$} & $\mathit{99.77}$ & $95.30$ & $99.84$ & $97.82$ & $95.88$ & $91.07$ & $\cellcolor{orange!10}96.61$ \\
    {} & \textbf{Structured\ $\rho{=}0.2$} & $98.17$ & $81.74$ & $98.01$ & $89.85$ & $90.63$ & $76.20$ & $89.10$ \\
    {} & \textbf{Structured\ $\rho{=}0.3$} & $82.90$ & $58.71$ & $79.10$ & $71.93$ & $68.22$ & $56.72$ & $69.60$ \\
    {} & \textbf{Structured\ $\rho{=}0.4$} & $73.83$ & $57.83$ & $73.36$ & $60.85$ & $60.67$ & $54.62$ & $63.53$ \\
    {} & \textbf{Structured\ $\rho{=}0.5$} & $72.47$ & $59.39$ & $71.50$ & $58.35$ & $58.92$ & $52.77$ & $62.23$ \\
    {} & \textbf{Structured\ $\rho{=}0.6$} & $65.63$ & $55.69$ & $69.29$ & $53.88$ & $56.70$ & $51.63$ & $58.80$ \\
    {} & \textbf{Structured\ $\rho{=}0.7$} & $51.12$ & $50.19$ & $51.26$ & $50.10$ & $50.17$ & $50.60$ & $50.57$ \\
    {} & \textbf{Structured\ $\rho{=}0.8$} & $50.72$ & $50.43$ & $51.31$ & $50.20$ & $50.12$ & $50.08$ & $50.48$ \\
    {} & \textbf{Structured\ $\rho{=}0.9$} & $50.00$ & $50.00$ & $50.00$ & $50.00$ & $50.00$ & $50.00$ & $50.00$ \\
    \bottomrule

    \Block{2-9}{\textbf{Pruning Criterion - Comparison between $L_1$ magnitude and random model pruning} $[\uparrow]$} \\
    \\
    {} & \textbf{Methods} & \textbf{LFW} & \textbf{CFP-FP} & \textbf{CFP-FF} & \textbf{AgeDB-30} & \textbf{CALFW} & \textbf{CPLFW} & $\overline{\text{Acc}}$ [$\uparrow$] \\
    \midrule
    \Block[tikz={pattern = {Lines[angle=-45, distance=1.5mm, line width=0.5mm]}, pattern color=gray!50}]{1-1} {} & \textbf{ResNet-100 (unpruned)} & $\mathbf{99.80}$ & $\mathit{96.67}$ & $\mathit{99.89}$ & $98.35$ & $\mathit{96.15}$ & $\mathbf{93.32}$ & $\cellcolor{gray!30}97.36$ \\
    \dashmidrule
    \Block[tikz={pattern = {Lines[angle=-45, distance=1.5mm, line width=0.5mm]}, pattern color=cyan!50}]{9-1} {} & \textbf{$L_1$ Magnitude\ $\rho{=}0.1$} & $\mathbf{99.80}$ & $\mathit{96.67}$ & $\mathit{99.89}$ & $\mathbf{98.43}$ & $\mathbf{96.17}$ & $\mathit{93.23}$ & $97.37$ \\
    {} & \textbf{$L_1$ Magnitude\ $\rho{=}0.2$} & $\mathbf{99.80}$ & $\mathbf{96.70}$ & $\mathit{99.89}$ & $\mathbf{98.43}$ & $\mathbf{96.17}$ & $\mathit{93.23}$ & $\cellcolor{cyan!10}97.37$ \\
    {} & \textbf{$L_1$ Magnitude\ $\rho{=}0.3$} & $\mathbf{99.80}$ & $96.59$ & $\mathit{99.89}$ & $\mathit{98.42}$ & $96.08$ & $93.17$ & $97.32$ \\
    {} & \textbf{$L_1$ Magnitude\ $\rho{=}0.4$} & $\mathbf{99.80}$ & $96.44$ & $\mathit{99.89}$ & $98.32$ & $96.08$ & $93.18$ & $97.29$ \\
    {} & \textbf{$L_1$ Magnitude\ $\rho{=}0.5$} & $\mathbf{99.80}$ & $96.36$ & $\mathbf{99.90}$ & $98.20$ & $96.00$ & $92.70$ & $97.16$ \\
    {} & \textbf{$L_1$ Magnitude\ $\rho{=}0.6$} & $\mathbf{99.80}$ & $95.79$ & $\mathbf{99.90}$ & $98.05$ & $96.02$ & $92.22$ & $96.96$ \\
    {} & \textbf{$L_1$ Magnitude\ $\rho{=}0.7$} & $99.75$ & $94.87$ & $99.79$ & $97.50$ & $95.85$ & $90.38$ & $96.36$ \\
    {} & \textbf{$L_1$ Magnitude\ $\rho{=}0.8$} & $99.57$ & $87.50$ & $99.39$ & $94.70$ & $94.35$ & $82.92$ & $93.07$ \\
    {} & \textbf{$L_1$ Magnitude\ $\rho{=}0.9$} & $90.45$ & $65.76$ & $91.87$ & $75.27$ & $78.37$ & $58.87$ & $76.76$ \\
    \dashmidrule
    \Block[tikz={pattern = {Hatch[angle=45, distance=1.5mm, line width=0.5mm]}, pattern color=violet!40}]{9-1} {} & \textbf{Random $\rho{=}0.1$} & $97.43$ & $77.86$ & $98.03$ & $82.82$ & $89.18$ & $73.52$ & $\cellcolor{violet!10}86.47$ \\
    {} & \textbf{Random $\rho{=}0.2$} & $77.00$ & $59.67$ & $75.30$ & $57.85$ & $62.23$ & $58.23$ & $65.05$ \\
    {} & \textbf{Random $\rho{=}0.3$} & $71.73$ & $59.83$ & $73.21$ & $56.63$ & $57.88$ & $55.07$ & $62.39$ \\
    {} & \textbf{Random $\rho{=}0.4$} & $64.80$ & $56.89$ & $71.83$ & $57.47$ & $56.75$ & $54.48$ & $60.37$ \\
    {} & \textbf{Random $\rho{=}0.5$} & $50.00$ & $50.00$ & $50.00$ & $50.00$ & $50.00$ & $50.00$ & $50.00$ \\
    {} & \textbf{Random $\rho{=}0.6$} & $50.00$ & $50.00$ & $50.00$ & $50.00$ & $50.00$ & $50.00$ & $50.00$ \\
    {} & \textbf{Random $\rho{=}0.7$} & $50.00$ & $50.00$ & $50.00$ & $50.00$ & $50.00$ & $50.00$ & $50.00$ \\
    {} & \textbf{Random $\rho{=}0.8$} & $50.00$ & $50.00$ & $50.00$ & $50.00$ & $50.00$ & $50.00$ & $50.00$ \\
    {} & \textbf{Random $\rho{=}0.9$} & $50.00$ & $50.00$ & $50.00$ & $50.00$ & $50.00$ & $50.00$ & $50.00$ \\
    \bottomrule
    \end{NiceTabular}
    }
    \label{tab:fr_pruning_comparison_full}
\end{table}

\begin{table}[!ht]
    \centering
\caption{Performance comparison of four FR models using pAUC scores (discard rate = 0.3, FMR = $10^{-4}$). The \textbf{best} and \textit{second-best} results per dataset are highlighted. The final column displays the average pAUC across all benchmarks. We exclude XQLFW from this average to prevent evaluation bias, as its quality labels were derived using SER-FIQ. The best average pAUC is highlighted in \begin{tabular}{c}\cellcolor{green!10}GREEN\end{tabular} for supervised approaches (marked using \textcolor{green!60}{green stripes}) , and \begin{tabular}{c}\cellcolor{blue!10}BLUE\end{tabular} for unsupervised approaches (marked with \textcolor{blue!60}{blue stripes}).}
    \resizebox{\columnwidth}{!}{%
\begin{NiceTabular}{c r |  r r r r r r r r | r}
\Block{2-11}{\textbf{ArcFace~\cite{deng2019arcface}} - $pAUC * 10^{3} \, ($FMR$=10^{-4}) \, [\downarrow]$} \\
 \\
 {} & \textbf{{Methods}} & \textbf{Adience} & \textbf{AgeDB-30} & \textbf{CFP-FP} & \textbf{LFW} & \textbf{CALFW} & \textbf{CPLFW} & \textbf{XQLFW} & \textbf{IJB-C} & $\overline{pAUC}$ \\
\midrule
\Block[tikz={pattern = {Lines[angle=-45, distance=1.0mm,  line width=0.5mm]},pattern color=green!40}]{9-1} {} & \textbf{RankIQ~\cite{RANKIQ_FIQA}} & $35.792$ & $17.131$ & $13.301$ & $0.929$ & $25.432$ & $48.441$ & $172.415$ & $12.200$ & $21.889$ \\
{} & \textbf{PFE~\cite{PFE_FIQA}} & $27.089$ & $12.675$ & $8.976$ & $0.921$ & $24.361$ & $42.758$ & $171.946$ & $11.003$ & $18.255$ \\
{} & \textbf{SDD-FIQA~\cite{SDDFIQA}} & $29.760$ & $10.189$ & $12.499$ & $0.963$ & $24.245$ & $41.998$ & $178.950$ & $11.039$ & $18.670$ \\
{} & \textbf{MagFace~\cite{MagFace}} & $27.522$ & $10.816$ & $7.495$ & $\mathbf{0.841}$ & $22.829$ & $39.988$ & $190.526$ & $10.883$ & $17.196$ \\
{} & \textbf{CR-FIQA(L)~\cite{boutros_2023_crfiqa}} & $\mathbf{22.352}$ & $9.983$ & $6.166$ & $1.012$ & $\mathbf{21.958}$ & $\mathbf{33.201}$ & $159.127$ & $10.114$ & $\cellcolor{green!10}14.970$ \\
{} & \textbf{DifFIQA(R)~\cite{10449044}} & $29.121$ & $13.729$ & $6.707$ & $0.930$ & $24.367$ & $33.550$ & $158.615$ & $\mathit{9.872}$ & $16.897$ \\
{} & \textbf{eDifFIQA(L)~\cite{babnikTBIOM2024}} & $25.487$ & $\mathbf{8.878}$ & $6.048$ & $0.908$ & $23.466$ & $\mathit{33.348}$ & $166.808$ & $\mathbf{9.790}$ & $15.418$ \\
{} & \textbf{CLIB-FIQA~\cite{Ou_2024_CVPR}} & $27.319$ & $\mathit{9.436}$ & $6.769$ & $0.915$ & $23.340$ & $33.627$ & $\mathbf{150.932}$ & $9.957$ & $15.909$ \\
{} & \textbf{ViT-FIQA(T)~\cite{atzori2025vitfiqaassessingfaceimage}} & $25.664$ & $10.734$ & $\mathbf{5.663}$ & $\mathit{0.896}$ & $23.614$ & $33.388$ & $\mathit{156.275}$ & $10.118$ & $15.725$ \\
\dashmidrule
\Block[tikz={preaction={fill, blue!40}, pattern = {Lines[angle=-45, distance=1.0mm,  line width=0.5mm]},pattern color=green!40}]{1-1} {} & \textbf{FROQ~\cite{FROQ}} & $31.630$ & $10.388$ & $8.296$ & $0.959$ & $23.453$ & $37.003$ & $167.705$ & $9.919$ & $17.378$ \\
\dashmidrule
\Block[tikz={pattern = {Lines[angle=-45, distance=1.0mm,  line width=0.5mm]},pattern color=blue!40}]{4-1} {} & \textbf{SER-FIQ~\cite{SERFIQ}} & $27.434$ & $12.283$ & $6.305$ & $0.975$ & $24.202$ & $35.086$ & $156.847$ & $10.093$ & $16.625$ \\
{} & \textbf{FaceQnet~\cite{hernandez2019faceqnet,faceqnetv1}} & $35.469$ & $12.704$ & $11.470$ & $1.132$ & $25.723$ & $65.278$ & $202.213$ & $12.698$ & $23.496$ \\
{} & \textbf{GraFIQs(L)~\cite{grafiqs}} & $23.757$ & $11.034$ & $7.103$ & $1.040$ & $23.900$ & $37.669$ & $158.682$ & $10.294$ & $\cellcolor{blue!10}16.399$ \\
{} & \textbf{ViTNT-FIQA~\cite{vitnt_fiqa}} & $25.196$ & $14.276$ & $7.340$ & $1.149$ & $24.402$ & $35.678$ & $158.107$ & $10.233$ & $16.896$ \\
\dashmidrule
\rowcolor{gray!10}
\Block[tikz={pattern = {Lines[angle=-45, distance=1.0mm,  line width=0.5mm]},pattern color=blue!40}]{1-1} {} & \textbf{PreFIQs (Ours)} & $\mathit{23.200}$ & $10.744$ & $\mathit{5.884}$ & $1.142$ & $\mathit{22.737}$ & $34.380$ & $160.838$ & $10.192$ & $\underline{15.469}$ \\
\bottomrule

\Block{2-11}{\textbf{CurricularFace~\cite{curricularFace}} - $pAUC * 10^{3} \, ($FMR$=10^{-4}) \, [\downarrow]$} \\
 \\
 {} & \textbf{{Methods}} & \textbf{Adience} & \textbf{AgeDB-30} & \textbf{CFP-FP} & \textbf{LFW} & \textbf{CALFW} & \textbf{CPLFW} & \textbf{XQLFW} & \textbf{IJB-C} & $\overline{pAUC}$ \\
\midrule
\Block[tikz={pattern = {Lines[angle=-45, distance=1.0mm,  line width=0.5mm]},pattern color=green!40}]{9-1} {} & \textbf{RankIQ~\cite{RANKIQ_FIQA}} & $28.973$ & $13.897$ & $12.347$ & $0.929$ & $23.731$ & $44.689$ & $152.020$ & $11.229$ & $19.399$ \\
{} & \textbf{PFE~\cite{PFE_FIQA}} & $22.063$ & $10.451$ & $8.741$ & $0.921$ & $23.196$ & $79.334$ & $\mathbf{137.743}$ & $10.231$ & $22.134$ \\
{} & \textbf{SDD-FIQA~\cite{SDDFIQA}} & $24.334$ & $11.549$ & $11.167$ & $0.963$ & $23.413$ & $44.873$ & $162.193$ & $10.053$ & $18.050$ \\
{} & \textbf{MagFace~\cite{MagFace}} & $22.276$ & $9.427$ & $7.489$ & $\mathbf{0.841}$ & $21.915$ & $38.775$ & $163.263$ & $9.987$ & $15.816$ \\
{} & \textbf{CR-FIQA(L)~\cite{boutros_2023_crfiqa}} & $21.058$ & $9.511$ & $5.964$ & $1.012$ & $\mathbf{21.397}$ & $29.961$ & $149.557$ & $9.247$ & $14.021$ \\
{} & \textbf{DifFIQA(R)~\cite{10449044}} & $23.109$ & $11.749$ & $5.982$ & $0.930$ & $22.762$ & $\mathit{29.538}$ & $141.513$ & $9.163$ & $14.747$ \\
{} & \textbf{eDifFIQA(L)~\cite{babnikTBIOM2024}} & $\mathit{20.309}$ & $\mathbf{8.948}$ & $\mathit{5.693}$ & $0.908$ & $21.994$ & $\mathbf{29.462}$ & $148.370$ & $\mathbf{9.044}$ & $\cellcolor{green!10}13.766$ \\
{} & \textbf{CLIB-FIQA~\cite{Ou_2024_CVPR}} & $21.731$ & $9.634$ & $6.076$ & $0.915$ & $21.897$ & $29.973$ & $141.835$ & $9.251$ & $14.211$ \\
{} & \textbf{ViT-FIQA(T)~\cite{atzori2025vitfiqaassessingfaceimage}} & $20.890$ & $10.593$ & $5.800$ & $\mathit{0.896}$ & $22.717$ & $29.590$ & $144.613$ & $9.322$ & $14.258$ \\
\dashmidrule
\Block[tikz={preaction={fill, blue!40}, pattern = {Lines[angle=-45, distance=1.0mm,  line width=0.5mm]},pattern color=green!40}]{1-1} {} & \textbf{FROQ~\cite{FROQ}} & $26.215$ & $10.810$ & $10.309$ & $0.959$ & $22.092$ & $33.024$ & $142.794$ & $\mathit{9.119}$ & $16.075$ \\
\dashmidrule
\Block[tikz={pattern = {Lines[angle=-45, distance=1.0mm,  line width=0.5mm]},pattern color=blue!40}]{4-1} {} & \textbf{SER-FIQ~\cite{SERFIQ}} & $23.264$ & $9.451$ & $5.808$ & $0.975$ & $22.653$ & $32.754$ & $\mathit{138.231}$ & $9.210$ & $\cellcolor{blue!10}14.874$ \\
{} & \textbf{FaceQnet~\cite{hernandez2019faceqnet,faceqnetv1}} & $29.768$ & $11.695$ & $11.412$ & $1.132$ & $24.303$ & $125.926$ & $175.338$ & $11.664$ & $30.843$ \\
{} & \textbf{GraFIQs(L)~\cite{grafiqs}} & $20.743$ & $9.199$ & $6.143$ & $1.040$ & $22.502$ & $47.312$ & $141.425$ & $9.456$ & $16.628$ \\
{} & \textbf{ViTNT-FIQA~\cite{vitnt_fiqa}} & $21.473$ & $12.199$ & $7.714$ & $1.149$ & $23.177$ & $31.136$ & $144.810$ & $9.464$ & $15.187$ \\
\dashmidrule
\rowcolor{gray!10}
\Block[tikz={pattern = {Lines[angle=-45, distance=1.0mm,  line width=0.5mm]},pattern color=blue!40}]{1-1} {} & \textbf{PreFIQs (Ours)} & $\mathbf{19.560}$ & $\mathit{9.020}$ & $\mathbf{5.533}$ & $1.142$ & $\mathit{21.648}$ & $32.238$ & $148.557$ & $9.425$ & $\underline{14.081}$ \\
\bottomrule

\Block{2-11}{\textbf{ElasticFace~\cite{elasticface}} - $pAUC * 10^{3} \, ($FMR$=10^{-4}) \, [\downarrow]$} \\
 \\
 {} & \textbf{{Methods}} & \textbf{Adience} & \textbf{AgeDB-30} & \textbf{CFP-FP} & \textbf{LFW} & \textbf{CALFW} & \textbf{CPLFW} & \textbf{XQLFW} & \textbf{IJB-C} & $\overline{pAUC}$ \\
\midrule
\Block[tikz={pattern = {Lines[angle=-45, distance=1.0mm,  line width=0.5mm]},pattern color=green!40}]{9-1} {} & \textbf{RankIQ~\cite{RANKIQ_FIQA}} & $32.314$ & $11.553$ & $9.822$ & $0.929$ & $22.184$ & $43.910$ & $153.491$ & $11.891$ & $18.943$ \\
{} & \textbf{PFE~\cite{PFE_FIQA}} & $23.534$ & $7.988$ & $7.174$ & $0.921$ & $22.110$ & $70.252$ & $160.004$ & $10.694$ & $20.382$ \\
{} & \textbf{SDD-FIQA~\cite{SDDFIQA}} & $26.484$ & $9.384$ & $7.541$ & $0.963$ & $22.013$ & $41.128$ & $185.308$ & $10.579$ & $16.870$ \\
{} & \textbf{MagFace~\cite{MagFace}} & $23.591$ & $7.355$ & $6.552$ & $\mathbf{0.841}$ & $20.984$ & $37.621$ & $170.809$ & $10.497$ & $15.349$ \\
{} & \textbf{CR-FIQA(L)~\cite{boutros_2023_crfiqa}} & $22.961$ & $7.895$ & $4.795$ & $1.012$ & $\mathit{20.747}$ & $29.283$ & $\mathit{145.298}$ & $9.764$ & $13.779$ \\
{} & \textbf{DifFIQA(R)~\cite{10449044}} & $25.311$ & $9.199$ & $4.956$ & $0.870$ & $21.605$ & $\mathbf{28.757}$ & $149.565$ & $\mathit{9.596}$ & $14.328$ \\
{} & \textbf{eDifFIQA(L)~\cite{babnikTBIOM2024}} & $23.121$ & $\mathbf{7.031}$ & $\mathbf{4.566}$ & $0.848$ & $20.828$ & $\mathit{28.759}$ & $161.211$ & $\mathbf{9.511}$ & $\cellcolor{green!10}13.523$ \\
{} & \textbf{CLIB-FIQA~\cite{Ou_2024_CVPR}} & $24.724$ & $7.561$ & $5.055$ & $\mathit{0.842}$ & $20.782$ & $29.149$ & $159.775$ & $9.701$ & $13.974$ \\
{} & \textbf{ViT-FIQA(T)~\cite{atzori2025vitfiqaassessingfaceimage}} & $\mathit{22.535}$ & $8.123$ & $4.803$ & $0.896$ & $21.569$ & $29.220$ & $172.150$ & $9.764$ & $13.844$ \\
\dashmidrule
\Block[tikz={preaction={fill, blue!40}, pattern = {Lines[angle=-45, distance=1.0mm,  line width=0.5mm]},pattern color=green!40}]{1-1} {} & \textbf{FROQ~\cite{FROQ}} & $28.572$ & $9.162$ & $7.599$ & $0.959$ & $\mathbf{20.723}$ & $32.372$ & $154.630$ & $9.615$ & $15.572$ \\
\dashmidrule
\Block[tikz={pattern = {Lines[angle=-45, distance=1.0mm,  line width=0.5mm]},pattern color=blue!40}]{4-1} {} & \textbf{SER-FIQ~\cite{SERFIQ}} & $25.769$ & $8.272$ & $4.833$ & $0.975$ & $21.701$ & $31.266$ & $\mathbf{143.390}$ & $9.659$ & $\cellcolor{blue!10}14.639$ \\
{} & \textbf{FaceQnet~\cite{hernandez2019faceqnet,faceqnetv1}} & $32.043$ & $9.529$ & $9.728$ & $1.059$ & $23.390$ & $112.090$ & $195.806$ & $12.382$ & $28.603$ \\
{} & \textbf{GraFIQs(L)~\cite{grafiqs}} & $22.707$ & $8.521$ & $5.193$ & $1.040$ & $21.409$ & $43.509$ & $177.210$ & $10.044$ & $16.060$ \\
{} & \textbf{ViTNT-FIQA~\cite{vitnt_fiqa}} & $23.405$ & $9.884$ & $5.827$ & $1.149$ & $22.154$ & $30.853$ & $166.858$ & $9.907$ & $14.740$ \\
\dashmidrule
\rowcolor{gray!10}
\Block[tikz={pattern = {Lines[angle=-45, distance=1.0mm,  line width=0.5mm]},pattern color=blue!40}]{1-1} {} & \textbf{PreFIQs (Ours)} & $\mathbf{20.954}$ & $\mathit{7.150}$ & $\mathit{4.711}$ & $1.142$ & $20.936$ & $30.677$ & $166.140$ & $9.964$ & $\underline{13.648}$ \\
\bottomrule

\Block{2-11}{\textbf{MagFace~\cite{MagFace}} - $pAUC * 10^{3} \, ($FMR$=10^{-4}) \, [\downarrow]$} \\
 \\
 {} & \textbf{{Methods}} & \textbf{Adience} & \textbf{AgeDB-30} & \textbf{CFP-FP} & \textbf{LFW} & \textbf{CALFW} & \textbf{CPLFW} & \textbf{XQLFW} & \textbf{IJB-C} & $\overline{pAUC}$ \\
\midrule
\Block[tikz={pattern = {Lines[angle=-45, distance=1.0mm,  line width=0.5mm]},pattern color=green!40}]{9-1} {} & \textbf{RankIQ~\cite{RANKIQ_FIQA}} & $35.315$ & $23.671$ & $20.646$ & $1.207$ & $23.341$ & $120.130$ & $178.752$ & $13.872$ & $34.026$ \\
{} & \textbf{PFE~\cite{PFE_FIQA}} & $26.848$ & $18.720$ & $9.904$ & $0.946$ & $22.728$ & $121.181$ & $178.028$ & $12.481$ & $30.401$ \\
{} & \textbf{SDD-FIQA~\cite{SDDFIQA}} & $29.644$ & $14.141$ & $14.142$ & $0.987$ & $22.965$ & $91.491$ & $196.530$ & $12.468$ & $26.548$ \\
{} & \textbf{MagFace~\cite{MagFace}} & $25.897$ & $14.706$ & $10.157$ & $\mathbf{0.865}$ & $21.743$ & $62.562$ & $190.811$ & $12.241$ & $21.167$ \\
{} & \textbf{CR-FIQA(L)~\cite{boutros_2023_crfiqa}} & $\mathit{23.460}$ & $13.942$ & $\mathbf{6.347}$ & $0.961$ & $21.650$ & $\mathbf{48.232}$ & $177.183$ & $11.418$ & $\cellcolor{green!10}18.001$ \\
{} & \textbf{DifFIQA(R)~\cite{10449044}} & $28.100$ & $19.892$ & $11.839$ & $0.990$ & $22.712$ & $63.122$ & $176.827$ & $\mathit{11.162}$ & $22.545$ \\
{} & \textbf{eDifFIQA(L)~\cite{babnikTBIOM2024}} & $25.601$ & $\mathbf{12.583}$ & $10.986$ & $1.003$ & $21.536$ & $62.819$ & $176.440$ & $\mathbf{11.076}$ & $20.801$ \\
{} & \textbf{CLIB-FIQA~\cite{Ou_2024_CVPR}} & $27.333$ & $\mathit{13.463}$ & $11.688$ & $0.993$ & $\mathit{21.473}$ & $63.054$ & $180.233$ & $11.278$ & $21.326$ \\
{} & \textbf{ViT-FIQA(T)~\cite{atzori2025vitfiqaassessingfaceimage}} & $25.507$ & $15.333$ & $\mathit{7.463}$ & $\mathit{0.938}$ & $22.271$ & $\mathit{48.300}$ & $176.617$ & $11.378$ & $18.741$ \\
\dashmidrule
\Block[tikz={preaction={fill, blue!40}, pattern = {Lines[angle=-45, distance=1.0mm,  line width=0.5mm]},pattern color=green!40}]{1-1} {} & \textbf{FROQ~\cite{FROQ}} & $32.243$ & $15.344$ & $10.360$ & $1.020$ & $21.775$ & $51.720$ & $\mathit{175.071}$ & $11.189$ & $20.522$ \\
\dashmidrule
\Block[tikz={pattern = {Lines[angle=-45, distance=1.0mm,  line width=0.5mm]},pattern color=blue!40}]{4-1} {} & \textbf{SER-FIQ~\cite{SERFIQ}} & $27.398$ & $18.478$ & $10.351$ & $1.343$ & $22.145$ & $57.925$ & $\mathbf{164.777}$ & $11.301$ & $21.277$ \\
{} & \textbf{FaceQnet~\cite{hernandez2019faceqnet,faceqnetv1}} & $34.071$ & $17.753$ & $18.253$ & $1.175$ & $23.538$ & $189.005$ & $201.488$ & $14.358$ & $42.593$ \\
{} & \textbf{GraFIQs(L)~\cite{grafiqs}} & $24.288$ & $14.670$ & $12.381$ & $1.280$ & $21.873$ & $72.939$ & $183.078$ & $11.785$ & $22.745$ \\
{} & \textbf{ViTNT-FIQA~\cite{vitnt_fiqa}} & $25.230$ & $21.339$ & $9.409$ & $1.116$ & $22.425$ & $50.261$ & $176.733$ & $11.507$ & $\cellcolor{blue!10}20.184$ \\
\dashmidrule
\rowcolor{gray!10}
\Block[tikz={pattern = {Lines[angle=-45, distance=1.0mm,  line width=0.5mm]},pattern color=blue!40}]{1-1} {} & \textbf{PreFIQs (Ours)} & $\mathbf{23.223}$ & $16.512$ & $11.027$ & $1.105$ & $\mathbf{21.322}$ & $56.892$ & $176.660$ & $11.559$ & $\underline{20.234}$ \\
\bottomrule
\end{NiceTabular}
}
\label{tab:sota_pauc_1e4}
\end{table}


\begin{table}[!ht]
    \centering
\caption{Performance of \textbf{ResNet50 trained on CASIA-Webface~\cite{casia_webface} using ArcFace~\cite{deng2019arcface}} on four FR models using pAUC scores (discard rate = 0.3, FMR = $10^{-3}$). 
ResNet-50 is pruned using unstructured $L_1$ magnitude pruning.
The \textbf{best} and \textit{second-best} results per dataset are highlighted. The final column displays the average pAUC across all benchmarks. We exclude XQLFW from this average, as its quality labels were derived using SER-FIQ.}
    \resizebox{\columnwidth}{!}{%
\begin{NiceTabular}{c r |  r r r r r r r | r}
\Block{2-10}{\textbf{ArcFace~\cite{deng2019arcface}} - $pAUC * 10^{3} \, ($FMR$=10^{-3}) \, [\downarrow]$} \\
 \\
 {} & \textbf{{Methods}} & \textbf{Adience} & \textbf{AgeDB-30} & \textbf{CFP-FP} & \textbf{LFW} & \textbf{CALFW} & \textbf{CPLFW} & \textbf{XQLFW} & $\overline{pAUC}$ \\
\midrule
\Block[tikz={pattern = {Lines[angle=-45, distance=1.5mm, line width=0.5mm]}, pattern color=cyan!50}]{9-1} {} & \textbf{Unstructured $\rho=$0.1} & $\mathit{13.045}$ & $\mathit{10.076}$ & $\mathbf{8.743}$ & $1.006$ & $22.172$ & $\mathbf{48.064}$ & $\mathit{187.383}$ & $\cellcolor{cyan!10}17.185$ \\
{} & \textbf{Unstructured $\rho=$0.2} & $13.636$ & $10.374$ & $9.044$ & $0.946$ & $22.022$ & $\mathit{49.660}$ & $188.583$ & $17.614$ \\
{} & \textbf{Unstructured $\rho=$0.3} & $13.174$ & $\mathbf{9.905}$ & $9.106$ & $0.933$ & $\mathit{21.853}$ & $54.262$ & $191.440$ & $18.205$ \\
{} & \textbf{Unstructured $\rho=$0.4} & $\mathbf{12.951}$ & $10.837$ & $\mathit{8.758}$ & $\mathit{0.874}$ & $\mathbf{21.769}$ & $51.052$ & $189.925$ & $17.707$ \\
{} & \textbf{Unstructured $\rho=$0.5} & $13.887$ & $10.856$ & $8.859$ & $\mathbf{0.798}$ & $22.416$ & $50.924$ & $192.805$ & $17.957$ \\
{} & \textbf{Unstructured $\rho=$0.6} & $13.572$ & $11.265$ & $8.937$ & $1.092$ & $22.836$ & $51.588$ & $193.798$ & $18.215$ \\
{} & \textbf{Unstructured $\rho=$0.7} & $13.677$ & $10.464$ & $9.877$ & $1.170$ & $22.623$ & $57.639$ & $194.791$ & $19.242$ \\
{} & \textbf{Unstructured $\rho=$0.8} & $14.668$ & $10.758$ & $12.219$ & $1.133$ & $23.531$ & $61.064$ & $189.744$ & $20.562$ \\
{} & \textbf{Unstructured $\rho=$0.9} & $15.590$ & $10.300$ & $12.350$ & $1.127$ & $22.830$ & $63.394$ & $\mathbf{186.573}$ & $20.932$ \\
\bottomrule

\Block{2-10}{\textbf{CurricularFace~\cite{curricularFace}} - $pAUC * 10^{3} \, ($FMR$=10^{-3}) \, [\downarrow]$} \\
 \\
 {} & \textbf{{Methods}} & \textbf{Adience} & \textbf{AgeDB-30} & \textbf{CFP-FP} & \textbf{LFW} & \textbf{CALFW} & \textbf{CPLFW} & \textbf{XQLFW} & $\overline{pAUC}$ \\
\midrule
\Block[tikz={pattern = {Lines[angle=-45, distance=1.5mm, line width=0.5mm]}, pattern color=cyan!50}]{9-1} {} & \textbf{Unstructured $\rho=$0.1} & $\mathbf{11.564}$ & $\mathit{10.730}$ & $9.258$ & $1.006$ & $21.376$ & $\mathbf{42.766}$ & $\mathit{168.681}$ & $16.116$ \\
{} & \textbf{Unstructured $\rho=$0.2} & $12.336$ & $\mathbf{10.176}$ & $\mathbf{8.556}$ & $1.001$ & $21.218$ & $\mathit{43.182}$ & $169.495$ & $\cellcolor{cyan!10}16.078$ \\
{} & \textbf{Unstructured $\rho=$0.3} & $11.935$ & $10.887$ & $9.140$ & $0.987$ & $\mathbf{20.985}$ & $44.248$ & $171.296$ & $16.364$ \\
{} & \textbf{Unstructured $\rho=$0.4} & $\mathit{11.659}$ & $11.070$ & $\mathit{8.867}$ & $\mathit{0.929}$ & $\mathit{21.081}$ & $43.385$ & $170.833$ & $16.165$ \\
{} & \textbf{Unstructured $\rho=$0.5} & $12.453$ & $10.971$ & $9.344$ & $\mathbf{0.818}$ & $21.500$ & $44.401$ & $172.527$ & $16.581$ \\
{} & \textbf{Unstructured $\rho=$0.6} & $12.117$ & $11.424$ & $9.312$ & $1.092$ & $21.901$ & $44.949$ & $172.224$ & $16.799$ \\
{} & \textbf{Unstructured $\rho=$0.7} & $12.250$ & $11.015$ & $9.771$ & $1.170$ & $21.788$ & $48.376$ & $172.332$ & $17.395$ \\
{} & \textbf{Unstructured $\rho=$0.8} & $12.980$ & $11.548$ & $12.407$ & $1.133$ & $22.365$ & $47.928$ & $169.064$ & $18.060$ \\
{} & \textbf{Unstructured $\rho=$0.9} & $13.900$ & $10.890$ & $13.347$ & $1.127$ & $21.893$ & $47.166$ & $\mathbf{165.039}$ & $18.054$ \\
\bottomrule

\Block{2-10}{\textbf{ElasticFace~\cite{elasticface}} - $pAUC * 10^{3} \, ($FMR$=10^{-3}) \, [\downarrow]$} \\
 \\
 {} & \textbf{{Methods}} & \textbf{Adience} & \textbf{AgeDB-30} & \textbf{CFP-FP} & \textbf{LFW} & \textbf{CALFW} & \textbf{CPLFW} & \textbf{XQLFW} & $\overline{pAUC}$ \\
\midrule
\Block[tikz={pattern = {Lines[angle=-45, distance=1.5mm, line width=0.5mm]}, pattern color=cyan!50}]{9-1} {} & \textbf{Unstructured $\rho=$0.1} & $\mathbf{14.466}$ & $\mathbf{9.271}$ & $8.451$ & $0.887$ & $21.282$ & $\mathbf{44.853}$ & $\mathit{176.500}$ & $\cellcolor{cyan!10}16.535$ \\
{} & \textbf{Unstructured $\rho=$0.2} & $15.203$ & $\mathit{9.298}$ & $\mathbf{7.565}$ & $0.827$ & $21.222$ & $45.541$ & $177.543$ & $16.609$ \\
{} & \textbf{Unstructured $\rho=$0.3} & $14.698$ & $9.927$ & $\mathit{7.782}$ & $0.814$ & $\mathbf{20.682}$ & $46.239$ & $179.343$ & $16.690$ \\
{} & \textbf{Unstructured $\rho=$0.4} & $\mathit{14.509}$ & $10.038$ & $7.969$ & $\mathit{0.755}$ & $\mathit{20.973}$ & $\mathit{45.307}$ & $178.396$ & $16.592$ \\
{} & \textbf{Unstructured $\rho=$0.5} & $15.394$ & $10.054$ & $8.633$ & $\mathbf{0.679}$ & $21.245$ & $46.766$ & $180.910$ & $17.129$ \\
{} & \textbf{Unstructured $\rho=$0.6} & $14.837$ & $10.378$ & $8.448$ & $0.974$ & $21.735$ & $47.489$ & $180.965$ & $17.310$ \\
{} & \textbf{Unstructured $\rho=$0.7} & $15.012$ & $9.671$ & $8.915$ & $1.053$ & $21.732$ & $50.233$ & $182.343$ & $17.769$ \\
{} & \textbf{Unstructured $\rho=$0.8} & $16.532$ & $9.672$ & $10.682$ & $1.017$ & $22.475$ & $53.326$ & $177.513$ & $18.951$ \\
{} & \textbf{Unstructured $\rho=$0.9} & $17.572$ & $9.808$ & $11.265$ & $1.043$ & $22.113$ & $54.480$ & $\mathbf{173.979}$ & $19.380$ \\
\bottomrule

\Block{2-10}{\textbf{MagFace~\cite{MagFace}} - $pAUC * 10^{3} \, ($FMR$=10^{-3}) \, [\downarrow]$} \\
 \\
 {} & \textbf{{Methods}} & \textbf{Adience} & \textbf{AgeDB-30} & \textbf{CFP-FP} & \textbf{LFW} & \textbf{CALFW} & \textbf{CPLFW} & \textbf{XQLFW} & $\overline{pAUC}$ \\
\midrule
\Block[tikz={pattern = {Lines[angle=-45, distance=1.5mm, line width=0.5mm]}, pattern color=cyan!50}]{9-1} {} & \textbf{Unstructured $\rho=$0.1} & $\mathbf{13.332}$ & $\mathbf{10.689}$ & $10.886$ & $1.032$ & $21.774$ & $\mathbf{52.677}$ & $\mathbf{196.726}$ & $\cellcolor{cyan!10}18.399$ \\
{} & \textbf{Unstructured $\rho=$0.2} & $14.026$ & $\mathit{10.845}$ & $\mathbf{10.483}$ & $\mathit{0.954}$ & $21.749$ & $\mathit{54.247}$ & $199.101$ & $18.717$ \\
{} & \textbf{Unstructured $\rho=$0.3} & $13.640$ & $10.886$ & $10.822$ & $0.977$ & $\mathbf{21.512}$ & $61.676$ & $202.518$ & $19.919$ \\
{} & \textbf{Unstructured $\rho=$0.4} & $\mathit{13.384}$ & $11.082$ & $\mathit{10.506}$ & $0.958$ & $\mathit{21.532}$ & $56.992$ & $199.355$ & $19.076$ \\
{} & \textbf{Unstructured $\rho=$0.5} & $14.278$ & $11.076$ & $10.957$ & $\mathbf{0.823}$ & $21.962$ & $55.992$ & $202.837$ & $19.181$ \\
{} & \textbf{Unstructured $\rho=$0.6} & $13.875$ & $11.975$ & $10.926$ & $1.118$ & $22.564$ & $57.054$ & $204.008$ & $19.585$ \\
{} & \textbf{Unstructured $\rho=$0.7} & $14.082$ & $11.216$ & $11.737$ & $1.239$ & $22.528$ & $77.373$ & $205.857$ & $23.029$ \\
{} & \textbf{Unstructured $\rho=$0.8} & $15.140$ & $11.339$ & $15.235$ & $1.250$ & $23.172$ & $86.173$ & $200.850$ & $25.385$ \\
{} & \textbf{Unstructured $\rho=$0.9} & $15.902$ & $11.143$ & $16.631$ & $1.298$ & $22.454$ & $92.730$ & $\mathit{197.217}$ & $26.693$ \\
\bottomrule
\end{NiceTabular}
}
\label{tab:resnet50_fiqa_pauc_fnmr1e-3}
\end{table}

\end{document}